\documentclass[preprint]{elsarticle}
\usepackage{graphicx} % Required for inserting images
\usepackage{enumitem}
\usepackage{amsmath}
\usepackage{amssymb}
\usepackage{caption}
\usepackage{subcaption}
\usepackage{hyperref}

\usepackage[a4paper, left=2.5cm, right=2.5cm, top=2.5cm, bottom=2.5cm]{geometry}

\newtheorem{prob}{Problem}
\newtheorem{rmrk}{Remark}

\newtheorem{assumption}{Assumption}

\begin{document}

\begin{frontmatter}

\title{Trajectory Tracking for Multi-Manipulator Systems in Constrained Environments\tnoteref{t1}}
\tnotetext[t1]{This work was supported by the ERC CoG LEAFHOUND, the Swedish Research Council (VR) and the Knut och Alice Wallenberg Foundation (KAW).}
\author[1]{Mayank Sewlia%
}
\ead{sewlia@kth.se}
\author[2]{Christos K. Verginis}
\ead{christos.verginis@angstrom.uu.se}
\author[1]{Dimos V. Dimarogonas}
\ead{dimos@kth.se}
\affiliation[1]{organization={Division of Decision and Control, Department of Electrical Engineering,
KTH Royal Institute of Technology},
city={Stockholm},
country={Sweden},
}
\affiliation[2]{organization={Signals and Systems, Department of Electrical Engineering},
addressline={Uppsala University},
city={Uppsala},
country={Sweden}}

\begin{abstract}
We consider the problem of cooperative manipulation by a mobile multi-manipulator system operating in obstacle-cluttered and highly constrained environments under spatio-temporal task specifications. The task requires transporting a grasped object while respecting both continuous robot dynamics and discrete geometric constraints arising from obstacles and narrow passages. To address this hybrid structure, we propose a multi-rate planning and control framework that combines offline generation of an STL-satisfying object trajectory and collision-free base footprints with online constrained inverse kinematics and continuous-time feedback control. The resulting closed-loop system enables coordinated reconfiguration of multiple manipulators while tracking the desired object motion. The approach is evaluated in high-fidelity physics simulations using three Franka Emika Panda mobile manipulators rigidly grasping an object.
\end{abstract}

\begin{abstract}
We consider the motion planning problem for a mobile multi-manipulator system grasping an object in an obstacle cluttered and highly constrained environment. The task is to move the object while satisfying spatio–temporal requirements expressed using Signal Temporal Logic (STL). Beyond avoiding static obstacles, the system must also navigate through narrow passages that induce discrete geometric ``modes'' requiring coordinated reconfiguration of all the manipulators, all while satisfying the task imposed on the object. This coupling of discrete transitions and continuous motions naturally gives rise to a hybrid systems problem. Our approach computes collision-free trajectories for both the grasped object and the mobile bases, followed by online inverse kinematics and a feedback controller to track these trajectories in real time. The resulting pipeline manages the hybrid interaction between high-level task specifications, discrete mode changes caused by spatial constraints, and the continuous dynamics of the mobile manipulators.
We demonstrate and validate our method in high-fidelity physics simulations using three Franka Emika Panda mobile manipulators rigidly grasping an object. %To our knowledge, these experiments provide one of the first demonstrations of a multi-manipulator system performing trajectory tracking through narrow passages under realistic hybrid constraints.
\end{abstract}

\end{frontmatter}
\section{Introduction}
A multi-manipulator system consists of multiple robotic arms sharing the load of carrying an object and transporting it to a desired location. When each robotic arm is mounted on a mobile base, the system is referred to as a mobile multi-manipulator system. Such systems are applicable in warehouse automation, automated construction, and search-and-rescue operations, where robots can autonomously navigate while carrying heavy payloads \cite{10.1108/IR-09-2014-0390}, \cite{nkwazema2021investigation}, \cite{Murphy2008}. These applications naturally involve hybrid interactions: the environment introduces discrete regions (free space, passages, tight corners), while the robots continuously try to adjust their configurations to avoid obstacles and maintain task requirements. 

The underlying task demands precise coordination among all mobile manipulators so as to avoid inducing unnecessary internal forces on the object, prevent self-collisions and inter-robot collisions, and ensure that the object satisfies its assigned spatio-temporal specification. The challenge becomes even greater in the presence of obstacles and narrow passages, which introduce discrete mode changes. For instance, if a passage is narrower than the convex hull of the combined system, the robots must transition into a different geometric configuration, reorienting and repositioning themselves without compromising the continuity of the object trajectory. This interplay between discrete structural changes in the environment and continuous control of the robots forms a natural hybrid systems setting.

This paper studies the problem of finding trajectories for all joints of the mobile multi-manipulator system such that the grasped object satisfies the STL defined task while the entire system avoids obstacles. Unlike previous works that focus primarily on moving around obstacles, we explicitly consider the problem of traversing narrow passages. These passages require coordinated reconfiguration of the entire manipulator system while still tracking the trajectory for the grasped object.
Our framework can be run in real time, enabling coordinated motion planning and control across multiple mobile manipulators while satisfying STL constraints on the grasped object. It consists of both offline and  online components. In the offline phase, we construct a map of the environment and compute a collision-free trajectory for the object using our recently developed MAPS$^2$ algorithm \cite{sewlia2023maps2}. MAPS$^2$ algorithm generates trajectories for multi-robot systems satisfying STL specifications. Next, we generate collision free trajectories for all mobile bases using a trajectory optimisation method. The online phase is divided into a slower, computationally intensive module and a faster feedback module. The slower module computes the joint angles of all manipulators given the object pose and base positions, while the faster module employs a feedback controller to track the desired joint angles. These modules still run online but at different frequencies allowing for realtime implementation. We demonstrate our approach in high-fidelity, physics-based simulations using three Panda mobile manipulators. These simulations highlight the ability of the framework to maintain safety, avoid collisions, and satisfy spatio-temporal specifications while traversing narrow passages.

\subsection{Related Work}
A variety of planning algorithms for mobile manipulators have been studied in the literature, with an excellent overview provided in \cite{machines10020097}. Broadly, motion planning for mobile manipulator systems can be categorised into three approaches: grid-based, sampling-based, and optimisation-based planners.

Grid-based planners, such as \cite{6225228}, \cite{6697080}, and \cite{5504833}, discretise the reachable space of the mobile manipulator and employ search-based methods to generate feasible trajectories as sequences of discretised regions. Other works, such as \cite{8593534}, construct inverse reachability maps to determine the joint configurations of the base and arms for a given end-effector location.

Sampling-based planners, on the other hand, rely on heuristics and approximations of the collision-free space. For example, \cite{7989390} proposes an asymptotically optimal manipulation planner by extending sampling-based roadmaps to the joint configuration space of both the manipulator and the object. Similarly, \cite{9662425} extends RRT to prehensible tasks, while \cite{7759547} introduces a bi-directional informed RRT* in which hyper-ellipsoidal regions are constructed for subsets of the configuration space of the mobile base and manipulators to iteratively refine trajectories. While sampling-based methods are highly effective in moderate dimensions (up to approximately 10), they scale poorly in cluttered environments, as the heuristics required to find collision-free paths become increasingly complex. Moreover, post-processing is often needed to enforce dynamical feasibility without compromising collision avoidance \cite{5152817}. Nonetheless, these methods retain desirable theoretical guarantees, such as probabilistic completeness and asymptotic optimality.

Optimisation-based planners often rely on sampling, either to generate good warm starts or within the solver itself, due to the nonlinear and nonconvex nature of the underlying problems. These methods are well suited to mobile manipulator systems, as they allow explicit enforcement of kino-dynamic and task-level constraints. The Graph of Convex Sets (GCS) framework \cite{Marcucci_2024} extends shortest-path search to hybrid spaces using convex sets and mixed-integer convex programming, while the Safe Boxes method \cite{10612232} builds on this idea by computing smooth, time-parameterised trajectories over precomputed convex cells. TrajOpt \cite{doi:10.1177/0278364914528132} formulates motion planning as sequential convex optimisation with signed-distance-based collision constraints, and CHOMP \cite{5152817} uses functional-gradient optimisation over continuous trajectories, though it is sensitive to initialisation and local minima. The GCS formulation becomes computationally slow even for dual-arm collision free motion, sometimes requiring up to 20 minutes to compute a trajectory. In contrast, TrajOpt and CHOMP are fundamentally single robot planners that require extensive customisation to handle a mobile multi-manipulator system, and they do not natively support closed kinematic chain constraints.

In the realm of multi-manipulator systems, recent work has advanced both control strategies and hybrid task–motion planning. For low-level control under contact uncertainty, \cite{10120601} proposes a distributed, event-triggered adaptive controller for teams manipulating with rolling contacts and unknown dynamics. Their scheme eliminates the need for force/torque sensing and guarantees slip avoidance through internal-force regulation. The work in \cite{verginis_timed} introduces timed abstractions for distributed cooperative manipulation: a model-free controller with prescribed performance and no inter-agent communication that lifts continuous dynamics to a finite timed transition system, thereby enabling the satisfaction of Metric-Interval Temporal Logic tasks over workspace partitions. At the planning level, \cite{zhang_geometric} address multi-robot geometric task-and-motion planning (MR-GTAMP) by combining motion-planner-derived reachability and occlusion information with a mixed-integer program to guide Monte Carlo Tree Search over collaborative action sequences. In \cite{robotics11060148}, the authors couple a completeness-guaranteed configuration-space cell decomposition with distributed low-level control (specifically, reference governors and prescribed-performance control) to achieve safety and convergence for teams transporting a rigidly grasped payload in clutter, without requiring continuous inter-robot communication. The closest work to our approach is \cite{patra2024kinodynamicmotionplanningcollaborative}, which formulates a kinodynamic motion planning framework for collaborative object transportation by multiple mobile manipulators in dynamic environments. Their method combines a global piecewise-linear path planner with a local receding-horizon trajectory generator that accounts for kinodynamic constraints, obstacle avoidance, and joint planning of both the bases and manipulators. However, obstacle avoidance is performed in 2D, where the robots are projected onto a plane, and the framework does not permit reconfiguration of either the manipulators or the bases. As a result, collisions may still occur in 3D, since the entire manipulator-base-object structure is treated as rigid.

\subsection{Outline}
%The rest of the paper is structured as follows: In Section \ref{sec:prob}, we present the necessary assumptions and state the problem we are solving, in Section \ref{sec:main} we present our main results discussing the proposed framework, while in Section \ref{sec:sim} we present the high-fidelity physics simulations and provide the conclusion in Section \ref{sec:conclusion}.
The remainder of the paper is structured as follows. Section~\ref{sec:prob} introduces the necessary assumptions and formally states the problem. Section~\ref{sec:main} presents the proposed framework and the main technical results. Section~\ref{sec:sim} presents high-fidelity physics simulations, and Section~\ref{sec:conclusion} concludes the paper.

\section{Problem Statement}\label{sec:prob}
We start by presenting the dynamics of the multi-manipulator system studied in this work and the environmental assumptions before formulating the problem.
\subsection{System Dynamics}
\begin{figure}
    \centering
    \includegraphics[width=\textwidth]{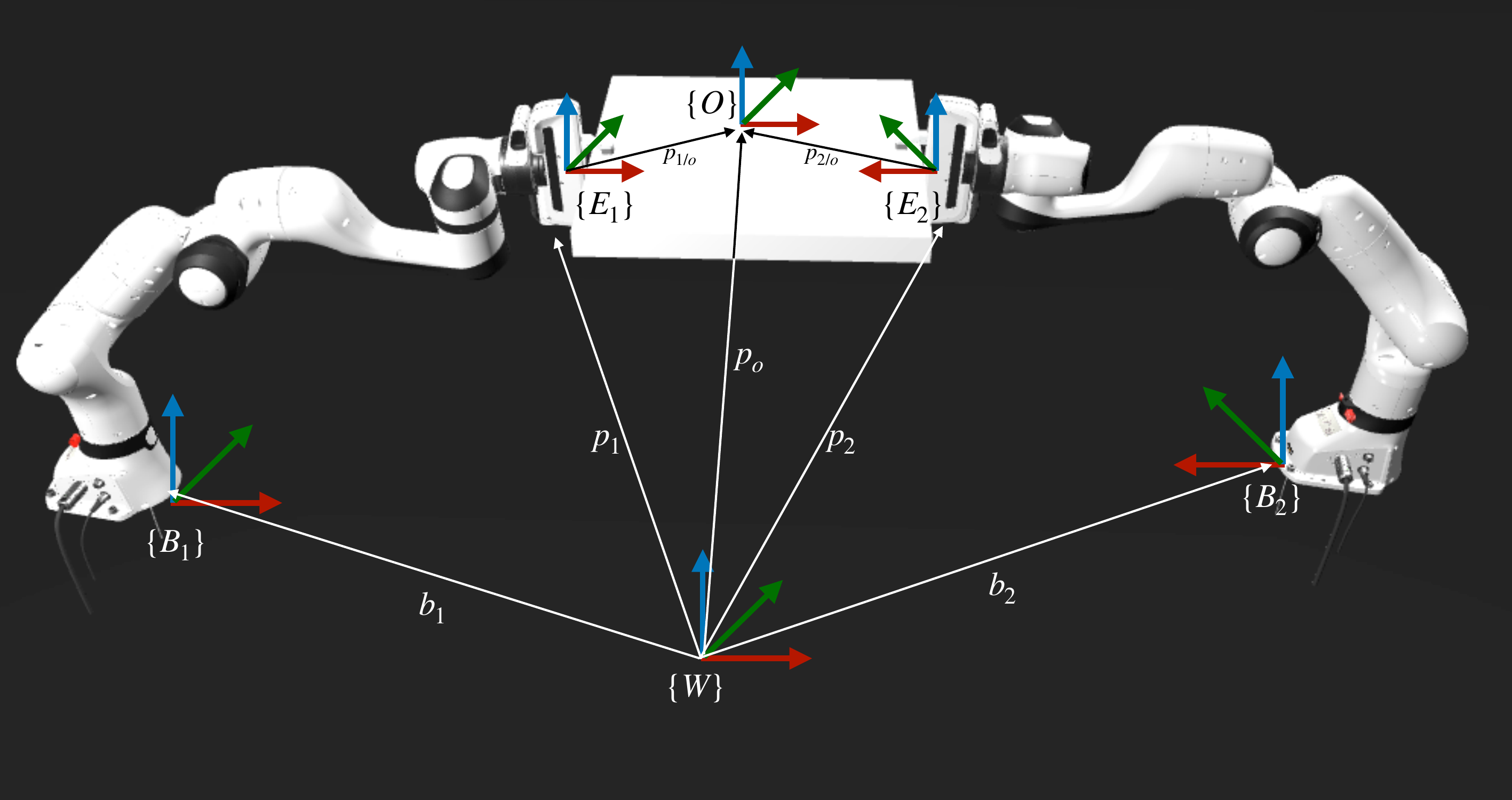}
    \caption{Visualisation of the frames for two Franka Emika Panda arms rigidly grasping an object \cite{haddadin2024franka}.}
    \label{fig_passage:frames}
\end{figure}

The system consists of $N$ mobile manipulators rigidly grasping an object. A rigid grasp does not allow relative motion between the end effectors and the grasped object. Let $\{W\}$ be the inertial (world) frame of reference. Each mobile manipulator is composed of a serial chain manipulator mounted on a mobile base. By $\{B_i\}$, we denote the frame attached to the mobile base of the $i$th mobile manipulator and by $\{E_i\}$ we denote the frame attached to the end-effector of the $i$th mobile manipulator, as shown in Figure \ref{fig_passage:frames}. The position of the base in the $\{W\}$ frame is denoted by $b_i\in\mathbb{R}^2$, and the position of the end-effector by $p_i\in\mathbb{R}^3$. The dynamics of each mobile manipulator is given by \cite{Murray561828},
\begin{equation}
    \label{eq_passage:agent_dynamics}
    M_i(q_i)\ddot{q}_i + C_i(q_i,\dot{q}_i) \dot{q}_i + g_i(q_i) = \tau_i
\end{equation}
where $q_i = \begin{bmatrix} b_i \\ \theta_i \end{bmatrix} \in \mathbb{R}^{n_i}$, with $b_i \in \mathbb{R}^2$ denoting the base position and $\theta_i \in \mathbb{R}^{n_i - 2}$ the joint angles of the arm. Here,  
$M_i(q_i)$ is the inertia matrix,  
$C_i(q_i,\dot{q}_i)\dot{q}_i$ captures Coriolis and centrifugal effects,  
$g_i(q_i)$ represents gravitational torques,  
and $\tau_i$ denotes the input control torques applied to the system.
%where $q_i = \begin{bmatrix}b_i \\ \theta_i\end{bmatrix}\in\mathbb{R}^{n_i},$
%with $b_i\in\mathbb{R}^2$ denotes the position of the base of the $i$th arm, and $\theta_i\in\mathbb{R}^{n_i-2}$ representing its joint angles. 
We currently present the formulation of a holonomic mobile base, where $b_i$ denotes the position in 2D and in Section \ref{footprint_planner} we show how this generalises to nonholonomic bases. The relationship between $q_i$ and $p_i$ is governed by the forward kinematics,
\begin{equation}
\label{eq_passage:forward_kinematics}
^WX_i^{E_i} = f(q_i)
\end{equation}
where $$^WX_i^{E_i} = \begin{bmatrix}^WR_i^{E_i} &p_i\\ \mathbf{0}_{1\times 3}&1\end{bmatrix}$$ represents the homogeneous transformation matrix from the $\{E_i\}$ frame to the $\{W\}$ frame, and $f(q_i):\mathbb{R}^{n_i}\to \mathrm{SE(3)}$ is a function that maps the joint configuration $q_i$ to the end-effector position $p_i$. 
The stacked dynamics of all mobile manipulators is given by
\begin{equation}
    \label{eq_passage:stacked_agent_dynamics}
    M(q)\ddot{q} + C(q,\dot{q}) \dot{q} + g(q) = \tau
\end{equation}
where, 
$M(q)=\text{diag}(M_1(q_1),\dots,$$M_N(q_N)),$
    $C(q,\dot{q})=\text{diag}(C_1(q_1,\dot{q}_1),\dots,$ \\$C_N(q_N,\dot{q}_N)),$
    $g(q)=\begin{bmatrix}g_1(q_1)^\top&\cdots&g_N(q_N)^\top\end{bmatrix}^\top,$ and, 
    $\tau=\begin{bmatrix}\tau_1^\top&\cdots&\tau_N^\top\end{bmatrix}^\top.
$
The pose and velocity of the object's center of mass are denoted by $x_o\in \mathbb{R}^6$  and $v_o\in\mathbb{R}^6$.
The dynamics of the object are
\begin{subequations}\label{eq_manip_intro:object_dynamics}
\begin{align}
    &\dot{{x}}_o=J_{o_r}^{-1}({x}_o)v_o, \\
    &M_o({x}_o)\dot{v}_o+C_o({x}_o,v_o)v_o+g_o({x}_o)=\lambda_o
\end{align}
\end{subequations}
where $M_o:\mathbb{R}^3\times \mathbb{T}^3\to\mathbb{R}^{6\times 6}$ is the positive-definite inertia matrix,  $C_o:\mathbb{R}^3\times \mathbb{T}^3\times \mathbb{R}^6\to \mathbb{R}^{6\times 6}$ is the Coriolis matrix and $g_o:\mathbb{R}^3\times \mathbb{T}^3\to \mathbb{R}^6$ is the gravity vector. Additionally,  $J_{o_r}:\mathbb{T}^3\to \mathbb{R}^{6\times6}$  is the object representation Jacobian and $\lambda_o\in\mathbb{R}^6$ is the force vector acting on the object's center of mass. 
When a robot interacts with an environment, it experiences forces as a result. The single robot dynamics in \eqref{eq_passage:agent_dynamics} in joint space then becomes in task space as 
\begin{equation}\label{agent_taskspace}
    \bar{M}(q_i)\dot{v_i}+\bar{C}(q_i,\dot{q}_i)v_i+\bar{g}(q_i)=u_i-\lambda_i,
\end{equation}
where $\bar{M}_i:\mathbb{R}^{n_i}\to \mathbb{R}^{6\times 6}$ is the task-space positive definite inertia matrix, $\bar{C}_i: \mathbb{R}^{2n_i}\to \mathbb{R}^{6\times 6}$ the task-space Coriolis matrix, $\bar{g}_i: \mathbb{R}^{n_i}\to \mathbb{R}^6$ the task-space gravity vector, $u_i\in \mathbb{R}^6$ is the task-space input wrench and $\lambda_i\in\mathbb{R}^6$ is the generalised force vector that agent $i$ exerts on the object.
\begin{assumption}
We assume that the robots grasp the object with rigid grasps, i.e., the relative pose between the end-effector frame of the \(i\)th robot, \(\{E_i\}\), and the object frame, \(\{O\}\), remains constant.
\end{assumption}

The rigid grasp assumption leads to the velocity constraint,
\[
J(q)\dot{q} = G^\top v_o
\]
where $J(q)=\text{diag}(J_i(q_i))$ is the block diagonal manipulator Jacobian.
The object dynamics \eqref{eq_manip_intro:object_dynamics} are coupled to the robot dynamics \eqref{agent_taskspace} by the grasping constraint \begin{equation}
    \label{eq_manip_intro:lambda_i}
\lambda_o=G\lambda,
\end{equation}
where $G:= G(q):  \mathbb{R}^{\sum_{i=1}^{N}n_i}\to$ $\mathbb{R}^{6\times 6N}$ is the full row-rank grasp matrix defined as
$$G(q)=\begin{bmatrix}J_{o_1}^\top(q_1)$
$&J_{o_2}^\top(q_2)&\cdots$
$&J_{o_N}^\top(q_N)\end{bmatrix}$$ where $J_{o_i}:\mathbb{R}^{n_i}\to\mathbb{R}^{6\times 6}$ is the object-to-$i$'th agent Jacobian
$
    J_{o_i}(q_i)=\begin{bmatrix}I_3& S(-p_{i/o}(q_i))\\ 0_{3\times 3} & I_3\end{bmatrix}.
$
 This results in the coupled object-robot dynamics
\begin{equation}
    \label{eq_passage:coupled_dynamics}
    \tilde{M}\dot{v}_o + \tilde{C}v_o + \tilde{g}  = G^\top u
\end{equation}
where
$
    \tilde{M} = M_o+G\bar{M}G^\top, 
    \tilde{C} = C_o+G\bar{M}\dot{G}^\top + GCG^\top,
    \tilde{g} = g_o+G\bar{g}.
$
 Observe that the position of the object can be obtained via the end effector position by the relation $^WR_i^{E_i}p_{i/o}+p_i$, see Figure \ref{fig_passage:frames}. However, we do not design $u$ from \eqref{eq_passage:coupled_dynamics} directly, but design $\tau$ from \eqref{eq_passage:stacked_agent_dynamics} and impose the grasping as constraints to an optimisation problem as seen in Section \ref{ik}. The task space control can be computed using,
$
\tau = J^\top G^{-\top}u
$. See \cite{verginis_timed} for more details on the coupled object-robot dynamics.

\subsection{Signal Temporal Logic}
The desired task for the object is specified using a Signal Temporal Logic (STL) formula over the centre of mass of the grasped object. 
Let ${x}:\mathbb{R}_{\geq 0}\to\mathbb{R}^n$ be a continuous-time signal. Signal Temporal Logic \cite{maler_monitoring_2004} is a predicate-based logic with the following syntax,
\begin{equation}\label{stl_general}
    \phi = \top \ |\ \mu\ |\ \neg \phi\ |\ \phi_1\land\phi_2\ |\ \phi_1\mathcal{U}_{[a,b]}\phi_2
\end{equation}
where $\phi_1,\phi_2$ are STL formulae and $\mu$ is a predicate of the form $\mu:\mathbb{R}^n\times\mathbb{R}_{\geq 0}\to \mathbb{B}$ defined via a predicate function $p:\mathbb{R}^n\times\mathbb{R}_{\geq 0}\to \mathbb{R}$ as
\begin{equation} \label{eq:mu}
    \mu=\begin{cases}\top & p({x},t)\geq 0\\ \bot & p({x},t)<0\end{cases}.
\end{equation}
{The satisfaction relation $({x},t)\models \phi$ indicates that signal ${x}$ satisfies $\phi$ at time $t$ and is defined recursively as follows:
\begin{alignat*}{2}
    &({x},t)\models \mu &&\Leftrightarrow  p({x}, t)\geq 0\\
    &({x},t)\models \neg \phi &&\Leftrightarrow  \neg(({x},t)\models \phi)\\
    &({x},t)\models \phi_1 \land \phi_2  &&\Leftrightarrow ({x},t)\models \phi_2 \land ({x},t)\models \phi_2\\
    & ({x},t)\models \phi_1 \mathcal{U}_{[a,b]}\phi_2 &&\Leftrightarrow \exists t_1\in[t+a, t+b] \text{ s.t. } ({x},t_1)\models \phi_2 \\ &&& \quad \land \forall t_2\in[t,t_1], ({x},t_2)\models \phi_1. 
\end{alignat*}
}
A signal ${x}$ satisfies the \textit{Until} operator, $({x},t)\models \phi_1\mathcal{U}_{[a,b]}\phi_2$, if $\phi_1$ holds at all times before $\phi_2$ holds and $\phi_2$ holds at some time instance between $a$ and $b$. 
More details on STL semantics can be found in \cite{lindemann2019control}.
For brevity, we omit further details on the STL formulation and instead rely on our recently developed MAPS$^2$ algorithm \cite{sewlia2023maps2} to generate the desired trajectory for the object from the given STL specification for a multi-robot system.

\subsection{Environment}\label{subsec:env}
Let $\mathcal{W}\subset\mathbb{R}^3$ denote the workspace within which the robots operate, and let $\mathcal{O}\subset\mathcal{W}$ represent the region occupied by static obstacles. We denote $\mathcal{S}_r$ as the set that consists of all points $p\in\mathcal{W}$ that physically belong to the coupled $N$ mobile manipulator system, i.e., they consist part of either the volume of the robots or the volume of the grasped object. Note that these points depend on the actual value of $q$. Let all points $o_p\in\mathcal{O}$ be the points that physically belong to all the obstacles; $o_p$ are assumed to be independent of time.  

\begin{assumption}\label{ass_passage:obs}
    The obstacle-free workspace $\mathcal{W}\setminus \mathcal{O}$ is non-empty, and the obstacle region $\mathcal{O}$ remains static over time.
\end{assumption}
This is a standard assumption that allows planning in the obstacle free region. It is computationally infeasible to evaluate the distance between all points \(p \in \mathcal S_r(q)\) and all points \(o_p \in \mathcal O\).  
Instead, we select a finite set of representative points from \(\mathcal S_r(q)\) that capture the geometric features of the robots and the grasped object, and a finite set of representative points on the surfaces of the obstacles \(\mathcal O\).  
For any such robot–obstacle representative pair, we define \(d_{r,o}(q)\in\mathbb{R}\) as the signed distance between the selected points: \(d_{r,o} > 0\) indicates separation, and \(d_{r,o} < 0\) indicates penetration. We further assume that these selected points faithfully detect collisions, in the sense that
\[
d_{r,o}(q) > 0 
\;\;\Longleftrightarrow\;\;
\mathcal S_r(q) \cap \mathcal O = \varnothing,
\qquad
d_{r,o}(q) \leq 0
\;\;\Longleftrightarrow\;\;
\mathcal S_r(q) \cap \mathcal O \neq \varnothing.
\]
In this work, the signed distance values are obtained directly from a physics engine.  
For real-world deployment, an RGB-D camera is typically used, and the resulting point cloud can be processed to construct a signed distance field.  
A representative example is NVIDIA's \texttt{nvblox} library, which uses the algorithm in \cite{10218983} to reconstruct a signed distance field from depth data.
Further, let
\[
\mathcal S_r(q) = \bigcup_{\ell \in \mathcal L} \mathcal S_\ell(q)
\]
be the union of the occupied volumes \(\mathcal S_\ell(q) \subset \mathcal{W}\) of rigid body $l$ belonging to the robots and the grasped object, indexed by \(\mathcal L\).
Let \(\mathcal C_{\mathrm{allow}} \subseteq \mathcal L \times \mathcal L\) denote the set of
pairs \((\ell_i,\ell_j)\) for which contact is allowed (e.g., between the gripper and the
object or any two adjacent links), and the rest as the disallowed pairs 
$
\mathcal C_{\mathrm{forbid}}
= \{(\ell_i,\ell_j) \in \mathcal L \times \mathcal L \mid i \neq j,\,
(\ell_i,\ell_j) \notin \mathcal C_{\mathrm{allow}}\}.
$
If
\begin{equation}\label{eq:selfcollision}
\mathcal S_{\ell_i}(q) \cap \mathcal S_{\ell_j}(q) = \varnothing
\qquad \forall\, (\ell_i,\ell_j) \in \mathcal C_{\mathrm{forbid}},
\end{equation}
then no self-collision occurs; that is, the robots do not collide with themselves,
with neighbouring robots, or with the manipulated object.
Note that generating an obstacle-free reference trajectory for the centre of mass of the object does not guarantee that the object will avoid collisions, nor does it guarantee that the mobile manipulators will avoid collisions with obstacles or with each other. From here on, for simplicity we further assume that $x_{o}\in\mathbb{R}^3$ and only captures the position of the object in 3D coordinates. 
The problem we address in this work is the following:

\begin{prob}
Consider a system of $N$ mobile manipulators rigidly grasping an object with coupled dynamics \eqref{eq_passage:coupled_dynamics}, and let an STL formula $\varphi$ of the form \eqref{stl_general} specify the desired spatio-temporal behaviour of the object state $x_o$. The objective is to design control torques $\tau_i$ in \eqref{eq_passage:agent_dynamics} for each mobile manipulator $i=1,\dots,N$ such that the resulting $(x_o,t)\models \varphi$, $d_{r,o}>0\ \forall t$ and \eqref{eq:selfcollision} holds. 
\end{prob}

\section{Main Results}\label{sec:main}

In this section, we present our main algorithm. We begin with an intuitive overview of the proposed approach. The method is illustrated in Figure \ref{fig_trajopt:block} and summarised below:
    \begin{figure}
    \centering
    \includegraphics[width=1\textwidth]{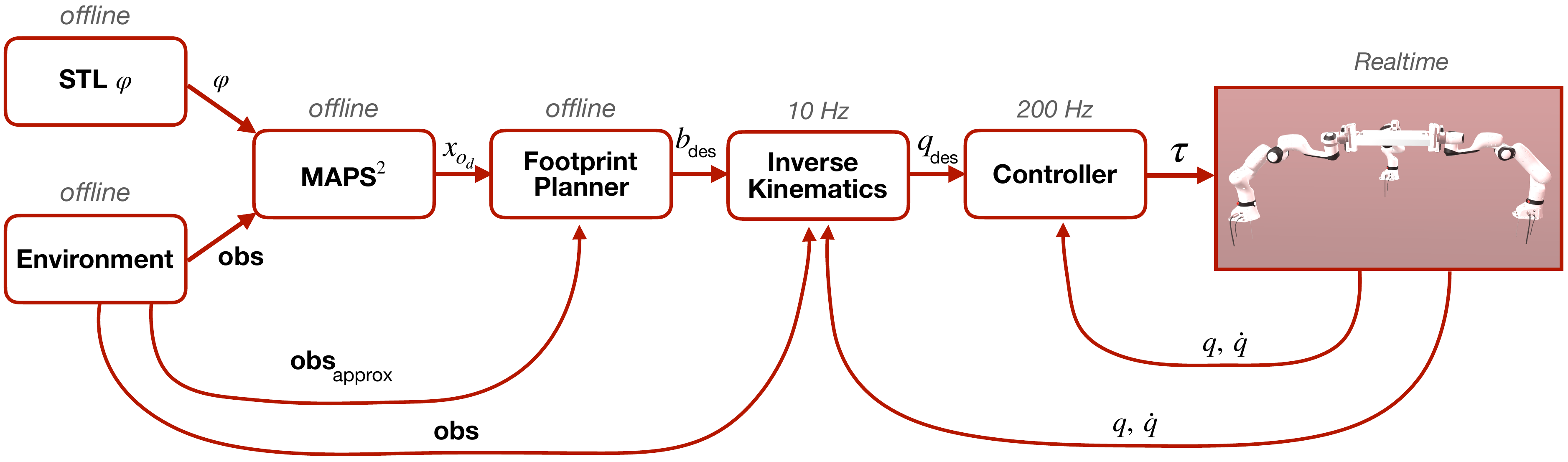}
    \caption{Proposed solution architecture.}
    \label{fig_trajopt:block}
\end{figure}%

\begin{enumerate}[label=(\Alph*)]
\item Compute a desired collision-free trajectory for the object $x_{o_d}$ using MAPS$^2$ (Section \ref{maps2}).
\item Determine desired collision-free base positions $b_{i\text{des}}$ for each mobile base $i$ using the \textit{Footprint Planner}, based on $x_{o_d}$ (Section \ref{footprint_planner}).
\item Compute the desired joint angles $q_{i\text{des}}$ using \textit{Inverse Kinematics}, given $b_{i\text{des}}$ and $x_{o_d}$ (Section \ref{ik}).
\item Design a control law to track the desired joint angles $q_{i\text{des}}$ for each agent $i=1,\dots,N$ (Section \ref{control}).
\end{enumerate}

Multi-robot manipulation in constrained environments naturally gives rise to a nonlinear hybrid systems problem. The workspace induces discrete geometric modes (e.g., free-space motion versus proximity to obstacles or narrow passages) that require coordinated reconfiguration of the entire team, while motion within each mode is governed by continuous robot and object dynamics. The proposed solution pipeline is hybrid and multi-rate: an offline layer generates an STL-satisfying object trajectory and a collision-free base footprint plan, while an online layer alternates between discrete inverse-kinematics updates at a slower rate and continuous-time feedback control at a faster rate using zero-order hold. This interaction between discrete mode changes, multi-rate computation, and continuous control places the problem and its solution squarely within the scope of nonlinear hybrid systems.

\subsection{MAPS$^2$:}\label{maps2} %Based on a given STL specification, the MAPS$^2$ algorithm computes a desired collision-free trajectory for the object $x_{o_d}(t)$. This desired trajectory is non-smooth as MAPS$^2$ is a sampling based planner. We post-process the trajectory using a two-step smoothing method. First, we interpolate the original waypoints using cubic splines to obtain a temporally dense and smooth trajectory. Then, we apply Gaussian smoothing to the interpolated positions to further reduce high-frequency noise. Finally, smoothed velocities are computed as time derivatives, and the result is represented as a Cubic Hermite Piecewise Polynomial for downstream use.

\begin{figure}
    \centering
    \includegraphics[width=1\textwidth]{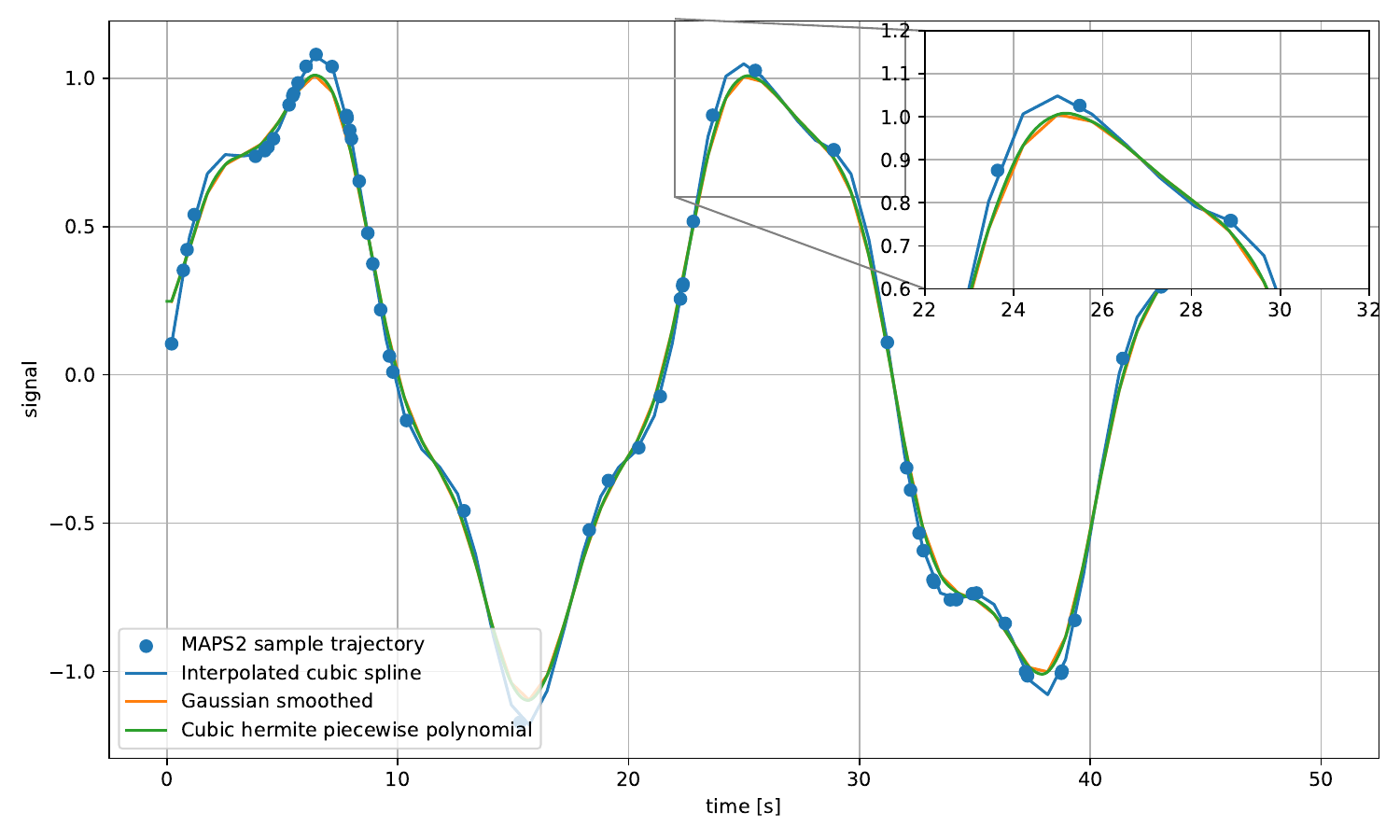}
    \caption{Post-processing of trajectories from MAPS$^2$.}
    \label{fig_trajopt:signals}
\end{figure}

Based on a given STL specification, the MAPS$^2$ algorithm computes a desired collision-free trajectory for the object, represented as a sequence of waypoints
\begin{equation}\label{maps_traj}
\{(t_k, x_{o_d}(t_k))\}_{k=0}^{K^{\prime}}, \qquad x_{o_d}(t_k)\in\mathbb{R}^3,
\end{equation}
where each waypoint corresponds to a sampled state for the object. As MAPS$^2$ is a sampling-based algorithm, the resulting trajectory is non-smooth. To address this, we perform a two-stage smoothing procedure that yields a smooth continuous trajectory. First, we apply cubic spline interpolation to each coordinate of the trajectory. Writing
$
x_{o_d}(t_k)=
\begin{bmatrix}
{}^x x_{o_d}(t_k)&
{}^y x_{o_d}(t_k)&
{}^z x_{o_d}(t_k)
\end{bmatrix}^\top,
$
we construct cubic spline interpolants
$\tilde{x}(t), \tilde{y}(t)$ and  $\tilde{z}(t)$
such that
$
\tilde{x}(t_k) = {}^x x_{o_d}(t_k),
\tilde{y}(t_k) = {}^y x_{o_d}(t_k),$ and  
$\tilde{z}(t_k) = {}^z x_{o_d}(t_k)$ for  $k=0,\dots,K^{\prime}.$
Since the $t_k$ are not evenly spaced as they are sampled randomly from a distribution, we construct a dense time grid,
$
\{t_j\}_{j=0}^{K}$ where $t_0 = t_{\min}$ and $t_K = t_{\max}$ are the minimum and maximum time instances from \eqref{maps_traj}. Next, we evaluate and interpolate the cubic spline on this dense time grid to obtain
$
\tilde{x}_{o_d}(t_j)=
\begin{bmatrix}
\tilde{x}(t_j)&
\tilde{y}(t_j)&
\tilde{z}(t_j)
\end{bmatrix}^\top 
$ where $ j=0,\dots,K.$
Spline interpolation can still preserve some oscillations from the original trajectory causing high velocities (local gradients). We therefore apply Gaussian smoothing to the interpolated sequence, implemented as a discrete convolution
$
x^{\mathrm{sm}}_{o_d}(t_j) \;=\;
\sum_{\ell=0}^{K} w_{j-\ell}\,\tilde{x}_{o_d}(t_\ell),$
$w_n \propto \exp\!\left(-\frac{n^2}{2\sigma^2}\right),
$
with the normalisation $\sum_n w_n = 1$. The parameter $\sigma>0$ controls the degree of smoothing. This produces smooth positions
$
x^{\mathrm{sm}}_{o_d}(t_j)=
\begin{bmatrix}
{}^x x^{\mathrm{sm}}_{o_d}(t_j)&
{}^y x^{\mathrm{sm}}_{o_d}(t_j)&
{}^z x^{\mathrm{sm}}_{o_d}(t_j)
\end{bmatrix}.
$ To obtain smooth differentiable trajectories for the solvers, $x^{\mathrm{sm}}_{o_d}(t_j)$ is encoded as a cubic Hermite piecewise polynomial \cite{APierceLecture} as follows,  \[
x^{\mathrm{ch}}_{o_d}(t) =
h_0(s)\,x^{\mathrm{sm}}_{o_d}(t_j)
+ h_1(s)\,x^{\mathrm{sm}}_{o_d}(t_{j+1})
+ h_2(s)\,\Delta t_j\, v_{o_d}(t_j)
+ h_3(s)\,\Delta t_j\, v_{o_d}(t_{j+1}),
\]
where
$
s = \frac{t - t_j}{\Delta t_j}, \Delta t_j = t_{j+1} - t_j,
$
and $h_0,h_1,h_2,h_3$ are the standard cubic Hermite basis functions
$
h_0(s) = 2s^3 - 3s^2 + 1,$
$h_1(s) = -2s^3 + 3s^2,$
$h_2(s) = s^3 - 2s^2 + s,$
and $h_3(s) = s^3 - s^2.
$ 
Here the velocities $v_{o_d}(t_j)$ were computed as 
$
v_{o_d}(t_j) =
\frac{x^{\mathrm{sm}}_{o_d}(t_{j+1}) - x^{\mathrm{sm}}_{o_d}(t_{j-1})}{t_{j+1} - t_{j-1}},
 j=1,\dots,K-1.
$
This construction yields a trajectory $x^{\mathrm{ch}}_{o_d}(t)\in\mathbb{R}^3$ that is smooth and continuously differentiable, thus suitable for downstream optimisation solvers. We henceforth abuse the notation and denote $x^{\mathrm{ch}}_{o_d}(t)$ as $x_{o_d}(t)$. These signals are depicted in the Figure \ref{fig_trajopt:signals}.

\subsection{Footprint Planner:} \label{footprint_planner}
Given the desired object trajectory $x_{o_d}(t)$, directly computing all joint angles $q_i$ for $i=1,\dots,N$ quickly becomes intractable, as it requires optimising over $N \times n_i$ decision variables and handling the non-convexities introduced by obstacles and forward kinematics. Moreover, optimising over a time horizon $[0, t_{\max}]$ with $K$ time steps results in $K \times N \times n_i$ decision variables. To mitigate this complexity, we decompose the planning problem into two stages: planning the mobile base trajectories and planning the manipulator arm configurations.
Given $x_{o_d}(t)$, the main objective is to compute a collision-free set of base trajectories
\[
b_{\mathrm{des}} =
\begin{bmatrix}
b_{1\mathrm{des}}^\top & \cdots & b_{N\mathrm{des}}^\top
\end{bmatrix}^\top
\]
that satisfy the following requirements:
\begin{itemize}
    \item The base trajectories of all mobile bases must not be too close such that the bases collide with each other.
    \item The bases must avoid collisions with obstacles.
    \item Finally, we need two anchor requirements. One that anchors bases close to the desired object trajectory $x_{o_d}(t)$ and another that anchors bases close to each other. The $b_{\mathrm{des}}$ that we compute here will only function as a reference for the inverse kinematics problem later on, thus instead of computing the exact forward kinematics we only provide soft thresholds for both these anchors.   
    % \item The trajectories must remain feasible with respect to the desired object trajectory $x_{o_d}(t)$, i.e., the bases must not drift so far apart that the manipulators cannot maintain the grasp close to $x_{o_d}(t)$. 
    % \item The trajectories must remain mutually feasible, ensuring that the agents do not drift apart to the extent that they release the object.
\end{itemize}

This problem can be formulated as a trajectory optimisation problem, where the objective is to find $b_{des}$ that satisfies the above constraints. Let the post-processed trajectory from MAPS$^2$ $x_{o_d}$ (as described in Section \ref{maps2}) be discretised into $K$ time steps over the horizon $[0,t_{\max}]$, where $t_{\max}$ is the time horizon of the STL formula (the final time instance of the trajectory $x_{o_d}$ from MAPS$^2$). The optimisation problem thus is as follows,
\begin{subequations} \label{eq_passage:opt}
    \begin{align}
        \min_{b_{des}} \quad&\sum_{k=1}^K\sum_{i=1}^N\sum_{\substack{j=1\\ i\neq j}}^N w_i \Big(\|b_{i\text{des}}[k]-b_{j\text{des}}[k]\|^2-\alpha_i^2\Big)^2 \label{line_passage:cost}\\
        \text{s.t.} \quad &\Big\|\frac{1}{N}\sum_{i=1}^N b_{i\text{des}}[k]-x_{o_d}(t)\Big\|^2\leq \epsilon \label{line_passage:centroid}\\ 
        & b_{i\text{des}}[k+1]-b_{i\text{des}}[k]\in [-\eta,\eta], \quad \forall\ i=1,\dots,N \label{line_passage:smooth}\\
        & z_{obj}[k] - {}^z x_{o_d}[k] \in [-\delta, \delta]\label{line_passage:object_height}\\
        & b_{\text{des}}[k] \in \mathcal{W}\setminus \mathcal{O}_{\text{approx}}\label{line_passage:obstacle}\\
        & b_{\text{des}}[0] = x_{o_d}(0)\label{line_passage:init}\\
        & b_{\text{des}}[t_{\max}] = x_{o_d}(t_{\max})\label{line_passage:final}
    \end{align}
\end{subequations}
Here, 
    \begin{itemize}
        \item \eqref{line_passage:cost} defines the cost function that penalises deviations of the squared distance between any two bases from a target squared distance $\alpha_i^2\ (\alpha_i\in\mathbb{R}_{>0})$. This cost essentially keeps the bases close to each other such that they do not drift too far apart and release teh object. The distance $\alpha_i$ is not uniform over all pairs of bases because if the manipulators grasp the object in a non-symmetric configuration (see Fig~\ref{fig_passage:panda3}, where robots~1 and~3 are farther apart than robots~1 and~2 or robots~2 and~3), then the values of $\alpha_i$ vary according to the kinematic reach from each base to its end-effector. The weights $w_i\in\mathbb{R}_{>0}$ are positive scalars that can be tuned to control how strictly the target inter-base distances are enforced. In this formulation, we consider all pairwise inter-base distances and penalise deviations from their desired values.
        \item \eqref{line_passage:centroid} ensures that the centroid of the base positions remains $\epsilon\in\mathbb{R}_{>0}$ close to the desired object trajectory $x_{o_d}$. Here, only the $x$ and $y$ coordinates of $x_{o_d}$ are considered, as the bases operate in a fixed height plane. This constraint maintains the proximity of the centroid formed by all base positions to the smoothened trajectory generated by the MAPS$^2$ algorithm, which already provides a globally feasible path - an advantage we aim to transfer to the base trajectories. 
        \item \eqref{line_passage:smooth} enforces a bounded step-to-step displacement, which corresponds to a velocity bound up to the discretisation factor. The allowable displacement $\eta\in\mathbb{R}_{>0}$ can depend on factors such as the object or payload weight, the friction properties of the floor, and the maximum admissible acceleration of the mobile bases. This constraint prevents abrupt motions between consecutive time steps and ensures that the base trajectories remain physically plausible and trackable.
        \item \eqref{line_passage:object_height} enforces a coupling between the spatial arrangement of the bases and the height of the grasped object. The average pairwise distance of the bases is computed at each time step and used to predict the object height via a simple linear model,
        $
            z_{\mathrm{obj}}[k] = z_{\mathrm{ref}} - \kappa\ \mathrm{spread}[k],
        $ where 
        $\mathrm{spread}[k] = \frac{2}{N(N-1)} \sum_{(i,j), i<j} \|b_{i\text{des}}[k]-b_{i\text{des}}[k]\|
        $, $z_{\mathrm{ref}}$ is a fixed distance and $\kappa$ is a scaling factor. 
        This predicted height is then required to remain within $\pm \delta$ of the desired object height obtained from the continuous MAPS$^2$ trajectory. This constraint ensures that the bases do not come too close to each other thereby raising the object over the desired object height, and also are not too far apart such that the object drops below the desired object height. 
        \item \eqref{line_passage:obstacle} represents the obstacle-avoidance constraint in two dimensions. The obstacle region $\mathcal{O}$ is assumed to be decomposed into convex components, and let
        \[
\mathcal O_{\mathrm{proj}}
:= \{\, (x,y) \in \mathbb R^2 \mid \exists\, z \in \mathbb R \text{ such that } (x,y,z) \in \mathcal O \,\}.
\]
This constraint ensures that the base positions remain collision free within the workspace. Each 2D obstacle is approximated by a super-ellipse in the $(x,y)$ plane. For each obstacle $\ell = 1,\dots,L$, let $c_\ell = (c^x_\ell, c^y_\ell)^\top$ denote its centre and let $a_\ell^x, a_\ell^y > 0$ denote its shape parameters along the $x$ and $y$ directions, respectively. We define the corresponding super-ellipse function as
        \[
            F_\ell(b)
            =
            \left(\frac{b_x - c^x_\ell}{a_\ell^x}\right)^{4}
            +
            \left(\frac{b_y - c^y_\ell}{a_\ell^y}\right)^{4},
            \qquad b = (b_x,b_y)^\top \in \mathbb{R}^2.
        \]
        The interior of obstacle~$\ell$ corresponds to points satisfying $F_\ell(b) < \rho_\ell$, and the feasible workspace is approximated by
        \[
            \mathcal{W} \setminus \mathcal{O}_{\text{approx}}
            =
            \big\{\, b \in \mathbb{R}^2 \;\big|\; F_\ell(b) \ge \rho_\ell,\ \forall\,\ell = 1,\dots,L \big\},
        \]
        where $\rho_\ell \ge 1$ specifies a safety margin around obstacle~$\ell$ and encapsulates the radii of the bases, and 
$
\mathcal O_{\mathrm{proj}} \subseteq \mathcal O_{\mathrm{approx}},
$
. Accordingly, the obstacle-avoidance constraint imposed on the base trajectories is
        \[
            F_\ell\big(b_{i\text{des}}[k]\big) \ge \rho_\ell,
            \qquad 
            i = 1,\dots,N,\ \ell = 1,\dots,L,\ k = 0,\dots,K,
        \]
        where $b_{i\text{des}}[k]$ denotes the planar position of base~$i$ at time index~$k$.
        \item \eqref{line_passage:init} and \eqref{line_passage:final} define the initial and final conditions for the base trajectories.
    \end{itemize}
Note that for nonholonomic mobile bases, the corresponding kinematic constraint can be incorporated as an additional equality constraint in the optimisation problem~\eqref{eq_passage:opt}.

    \subsection{Inverse Kinematics:}\label{ik} 
    
Given the desired object trajectory $x_{o_d}(t)$ and the base trajectories $b_{\mathrm{des}}(t)$ computed by the footprint planner, the goal of this stage is to compute a feasible joint configuration $q_{\mathrm{des}}(t)$ for all manipulators that realises a rigid multi-arm grasp of the object while ensuring kinematic constraints and obstacle avoidance. We denote the base trajectories $b_{\mathrm{des}}(t)$ as $\bar{b}_{\text{des}}$ in this section as the inverse kinematics will compute new positions for the bases and the $b_{\mathrm{des}}(t)$ from the footprint planner will be used as a soft constraint. Recall that the \textit{Footprint Planner} is an offline algorithm while the IK is an online algorithm. In the implementation, this is achieved by solving a inverse-kinematics (IK) problem at a certain frequency. The following IK problem is solved upto a fixed number of optimisation steps in order to ensure computation at a certain frequency. The considerations encoded in the IK formulation are summarised below.

%The desired base trajectories $b_{\text{des}}$(denoted as $\bar{b}_{\text{des}}$ here) enable the computation of desired joint angles $q_{\text{des}}=\begin{bmatrix}q_{1\text{des}}^\top & \cdots & b_{N\text{des}}^\top\end{bmatrix}^\top$. The considerations are as follows:
    \begin{itemize}
        \item The arms along with the grasped object should avoid collisions with obstacle, i.e., $d_{r,o}>0$.
        \item The arms should ensure tracking of the desired base trajectories $b_{i\text{des}}$ and the desired object trajectory $x_{o_d}$.
        \item The arms should maintain configurations that avoid kinematic singularities.
        \item The end-effectors should satisfy the rigid grasp constraints on the object.
        \item The solution must respect the joint limit constraints.
    \end{itemize}
    These considerations can be formulated as an optimisation problem, where the objective is to find the desired joint angles $q_{\text{des}}=\begin{bmatrix}q_{1\text{des}}^\top & \cdots & b_{N\text{des}}^\top\end{bmatrix}^\top$ that satisfy the above requirements. %Let the trajectory $q_{\text{des}}$ be discretised into $T$ time steps over the horizon $[0,t_f]$.
    \begin{subequations} \label{eq_passage:ik}
        \begin{align}
            \min_{q_{\text{des}}} \quad& \|q_{\text{des}}[k]-q_{0}\|^2 + \sum_{i=1}^N \Big\|\Big(\ ^WR_i^{E_i}p_{i/o}+p_i\Big)[k]-x_{o_d}(t)\Big\|^2 + c_{\text{collision}}\label{line_passage:ikcost}\\ 
            \text{s.t.} \quad & \ p_{i/o}[k] = p_{i/o}[0]\label{line_passage:rigidp}\\
            & ^OR_i^{E_i}[k] = ^OR_i^{E_i}[0]\label{line_passage:rigidr}\\ 
            & \bar{b}_{\text{des}}[k] -\nu \leq b_{\text{des}}[k] \leq \bar{b}_{\text{des}}[k] + \nu \label{line_passage:blim}\\
            & ^WX_i^{E_i}[k] = f(q_{i\text{des}}[k]) \label{line_passage:fk}
        \end{align} 
    \end{subequations}

\begin{itemize}
\item \eqref{line_passage:ikcost} defines the overall cost function, composed of three complementary terms. The first term penalises deviations of the joint angles from a reference configuration $q_0$, which is selected to promote well-conditioned, collision-free postures and to provide a regularising effect that improves numerical robustness of the IK solver. The second term promotes accurate tracking of the desired object trajectory by penalising the discrepancy between the object position, expressed as $({}^W\!R_i^{E_i}p_{i/o} + p_i)$, and the desired centre-of-mass trajectory $x_{o_d}(t)$. 

 The final term, $c_{\text{collision}}$, encodes obstacle avoidance penalty at the IK level using a smooth, signed distance based cost function. Let $d_{r,o}(q_{\text{des}})$ (as defined in Section \ref{subsec:env}) denote the signed distance between a robot geometry and obstacle geometry. %, and let $\nabla d_{r,o}(q_{\text{des}})$ be its gradient with respect to the configuration $q$, obtained via translational Jacobians. 
For each robot-obstacle representative pair, we define a local potential
\[
    \phi\bigl(d_{r,o}(q_{\text{des}})\bigr)
    = w_{r,o}(d_{r,o}(q_{\text{des}}))\,\exp \bigl(-\alpha\,(d_{r,o}(q_{\text{des}})-d_{\mathrm{safe}})^2\bigr),
\]
where $\alpha>0$ is a shaping parameter, $d_{\mathrm{safe}}>0$ is a prescribed safety margin, and $w_{r,o}$ is a distance-dependent weight that down-weights pairs that are already far from contact. The total collision cost is then obtained by summing these potentials over all robot-obstacle pairs,
\[
    c_{\text{collision}}(q)
    = \sum_{(r,o)} \phi\bigl(d_{r,o}(q_{\text{des}})\bigr).
\]
% with gradients computed analytically via the chain rule using $\nabla d_{r,o}(q_{\text{des}})$. 
This construction was chosen after extensive experimentation to satisfy three requirements simultaneously: (i) negligible influence when the manipulators are well separated from obstacles, (ii) rapidly increasing cost and informative gradients as the distance approaches the safety margin, and (iii) smoothness and differentiability with respect to $q_{\text{des}}$, which are crucial for reliable convergence of the nonlinear optimisation.

    \item \eqref{line_passage:rigidp} and \eqref{line_passage:rigidr} together impose the rigid-grasp constraints necessary for cooperative manipulation. Constraint~\ref{line_passage:rigidp} enforces a fixed relative position between the end-effector and its corresponding grasp point on the object, ensuring that $p_{i/o}$ does not drift over time. Constraint~\eqref{line_passage:rigidr} ensures that the relative orientation between the end-effector and the object remains constant, thereby enforcing a rigid grasp throughout the motion.

    \item \eqref{line_passage:blim} restricts the base positions to remain within a bounded neighbourhood of the desired base trajectory $\bar{b}_{\mathrm{des}}[k]$. The tolerance $\nu\geq 0$ provides the flexibility needed to accommodate minor adjustments for collision avoidance or kinematic feasibility, while still ensuring that the mobile bases track their planned motion closely.

    \item \eqref{line_passage:fk} enforces consistency between the decision variables and the robot's kinematics. This constraint ensures that the end-effector pose ${}^W\!X_i^{E_i}[k]$ corresponds to the forward kinematics $f(q_{i\mathrm{des}}[k])$ of the manipulator at the desired joint configuration. It guarantees that all position and orientation expressions appearing in the optimisation are derived from physically realisable robot configurations.
\end{itemize}

\begin{rmrk}
The IK optimisation problem~\eqref{eq_passage:ik} is not solved to full convergence but instead executed for a fixed number of iterations. This practice is common in nonlinear, nonconvex problems, where performing only a limited number of gradient-based steps yields a solution that is sufficiently close to a local optimum while significantly reducing computation time. The permissible iteration budget is ultimately dictated by the computational capabilities of the hardware on which the optimisation runs.
\end{rmrk}

\subsection{Control Design}\label{control}

The \textit{Inverse Kinematics} module provides the desired joint configurations $q_{\text{des}}$ at discrete update times $t_k$, based on the trajectories generated by \textit{MAPS$^2$} and the \textit{Footprint Planner}. Between updates, the desired configuration is held constant via a zero-order hold. Each manipulator then tracks its corresponding desired configuration using a standard joint-space Proportional–Derivative (PD) controller with gravity compensation:
\begin{equation}\label{eq_passage:control}
    \tau_i(t)
    = -k_p\bigl(q_i(t) - q_{i\text{des}}[k]\bigr)
      -k_v\,\dot{q}_i(t)
      + g_i(q_i(t)),
    \qquad t \in [t_k, t_{k+1}),
\end{equation}
where $k_p, k_v > 0$ are tunable gains, and $g_i(q_i)$ denotes the gravity compensation term from \eqref{eq_passage:agent_dynamics}. %Since the IK module provides quasi-static desired joint configurations without desired joint velocities, the controller implicitly regulates $\dot{q}_i(t)$ toward zero while driving $q_i(t)$ toward $q_{i\text{des}}[k]$. 
Let \[
e_i(t) := q_i(t) - q_{i,\mathrm{des}}[k],
\qquad
\dot e_i(t) := \dot q_i(t).
\]
Consider the Lyapunov function
$
V_i(e_i,\dot e_i)
= \tfrac12 \dot e_i^\top M(q_i)\dot e_i
  + \tfrac12 e_i^\top K_p e_i.
$
and the dynamics \eqref{eq_passage:agent_dynamics} with disturbances $w_i(t)$ that enter through the input channel as,
\begin{equation}
    M_i(q_i)\ddot{q}_i + C_i(q_i,\dot{q}_i) \dot{q}_i + g_i(q_i) = \tau_i + w_i
\end{equation}
Substituting \eqref{eq_passage:control} in the above equation, its derivative satisfies
\[
\dot V_i
\le - \dot e_i^\top K_v \dot e_i + \dot e_i^\top w_i(t).
\]
In the disturbance-free case $w_i(t)\equiv 0$ we obtain
\[
\dot V_i \le -\lambda_{\min}(K_v)\,\|\dot e_i\|^2,
\]
where $\lambda_{\min}$ is the smallest eigenvalue of $K_c$, so the equilibrium $(e_i,\dot e_i)=(0,0)$ is globally exponentially stable. Moreover, when
$w_i(t)$ is bounded, this inequality yields an input--to--state stability
(ISS) bound for $(e_i,\dot e_i)$ with respect to $w_i(t)$, so the tracking
error remains ultimately bounded in the presence of moderate disturbances.
% If $w_i(t)$ is bounded, $\|w_i(t)\|\le \bar w_i$, then
% \[
% \dot V_i
% \le -\alpha \bigl(\|e_i\|^2+\|\dot e_i\|^2\bigr) + \beta \bar w_i^2
% \]
% for some $\alpha,\beta>0$, implying input-to-state stability and that the
% tracking error $(e_i,\dot e_i)$ remains ultimately bounded in the presence of
% moderate modelling errors and external disturbances.
% \section{Simulations}
This decentralised joint-space PD structure is sufficient in our setting, as the planning pipeline enforces multi-arm coordination and collision avoidance at the kinematic level, leaving the low-level controller to focus solely on accurate trajectory tracking.

\begin{figure}
    \centering
    \includegraphics[width=0.8\textwidth]{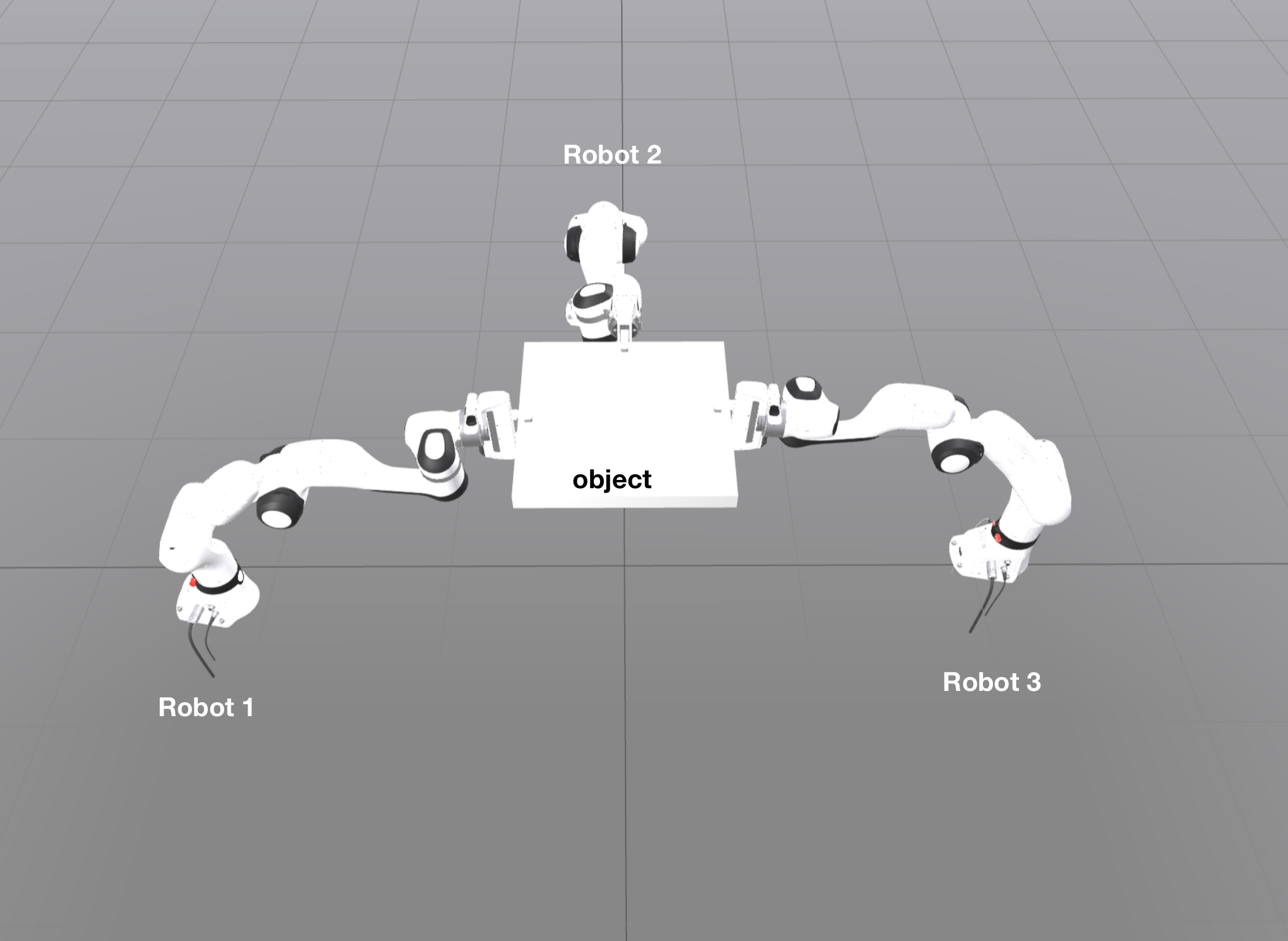}
    \caption{Setup of 3 panda arms developed by Franka Emika grabbing an object.}
    \label{fig_passage:panda3}
\end{figure}%
\begin{figure}
    \centering
    \includegraphics[width=0.8\textwidth]{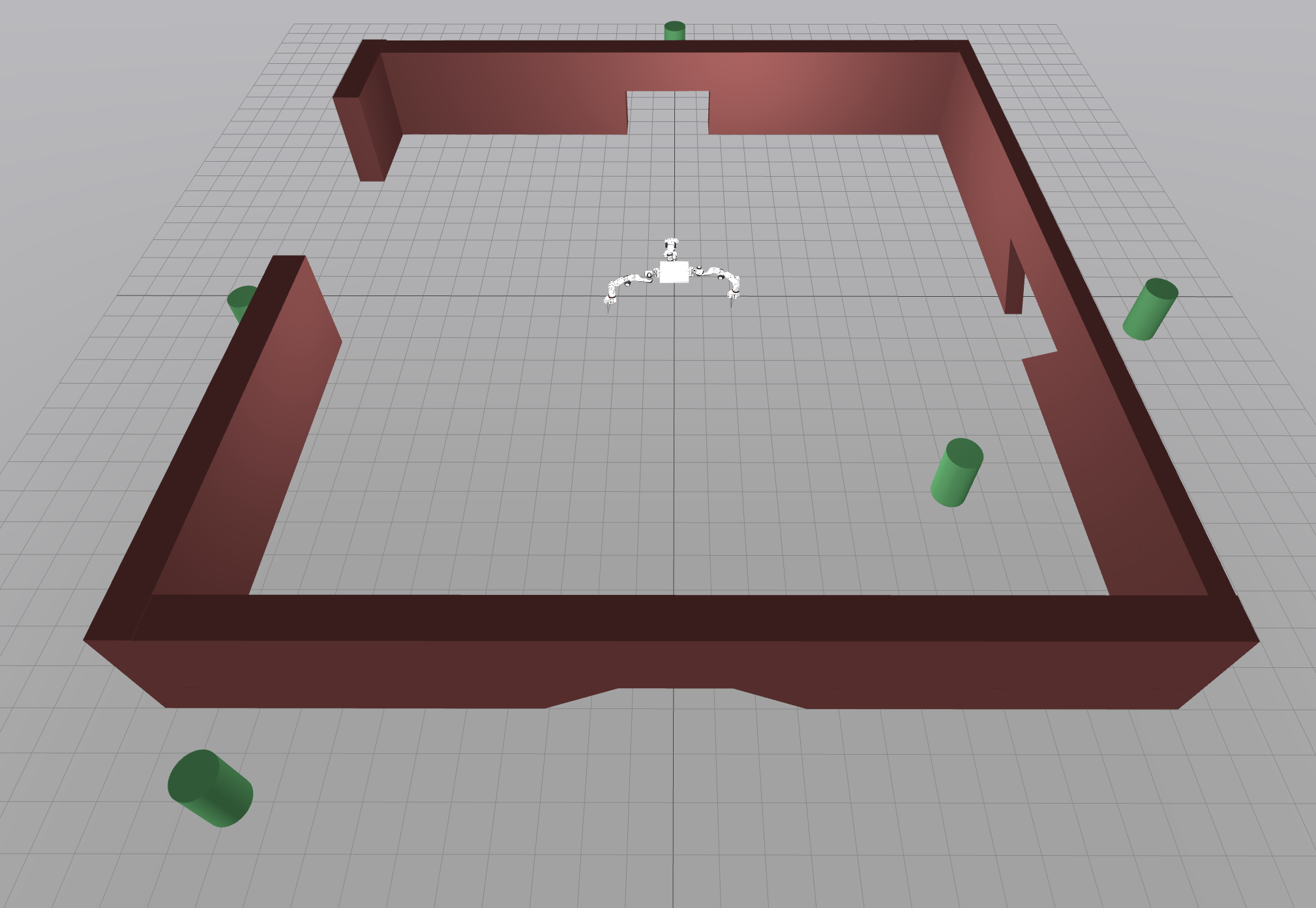}
    \caption{The workspace with obstacles shown in red, desired waypoints in green, and 3 panda mobile robots in white.}    
    \label{fig_passage:ws}
\end{figure}%
\begin{figure*}[t]
    \centering
    \begin{subfigure}[b]{0.49\textwidth}
        \centering
        \includegraphics[height=5.8cm]{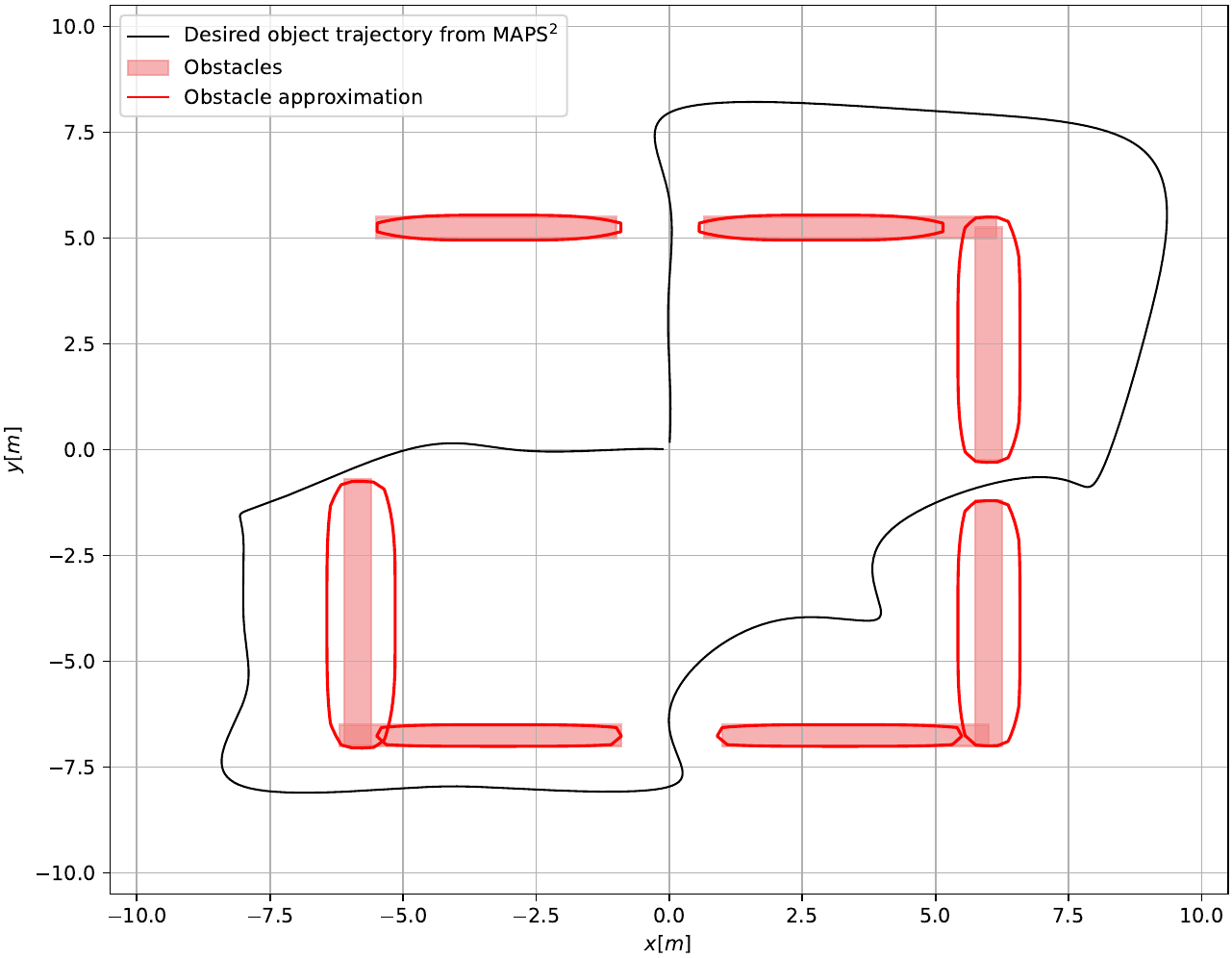}
    \end{subfigure}
    \hfill
    \begin{subfigure}[b]{0.49\textwidth}
        \includegraphics[height=5.8cm]{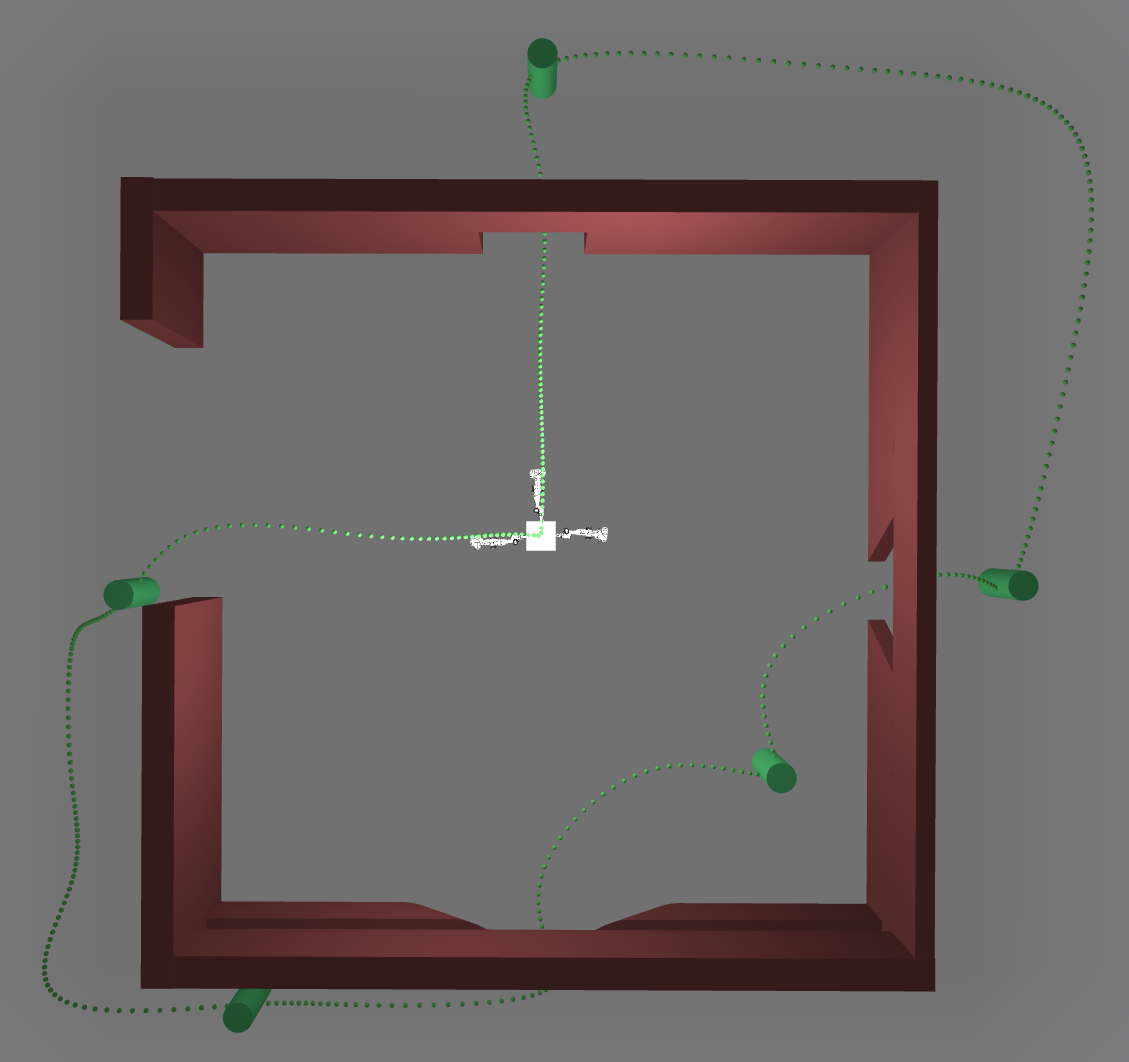}
    \end{subfigure}
    \vskip\baselineskip
    \begin{subfigure}[b]{0.49\textwidth}
        \centering
        \includegraphics[height=5.8cm]{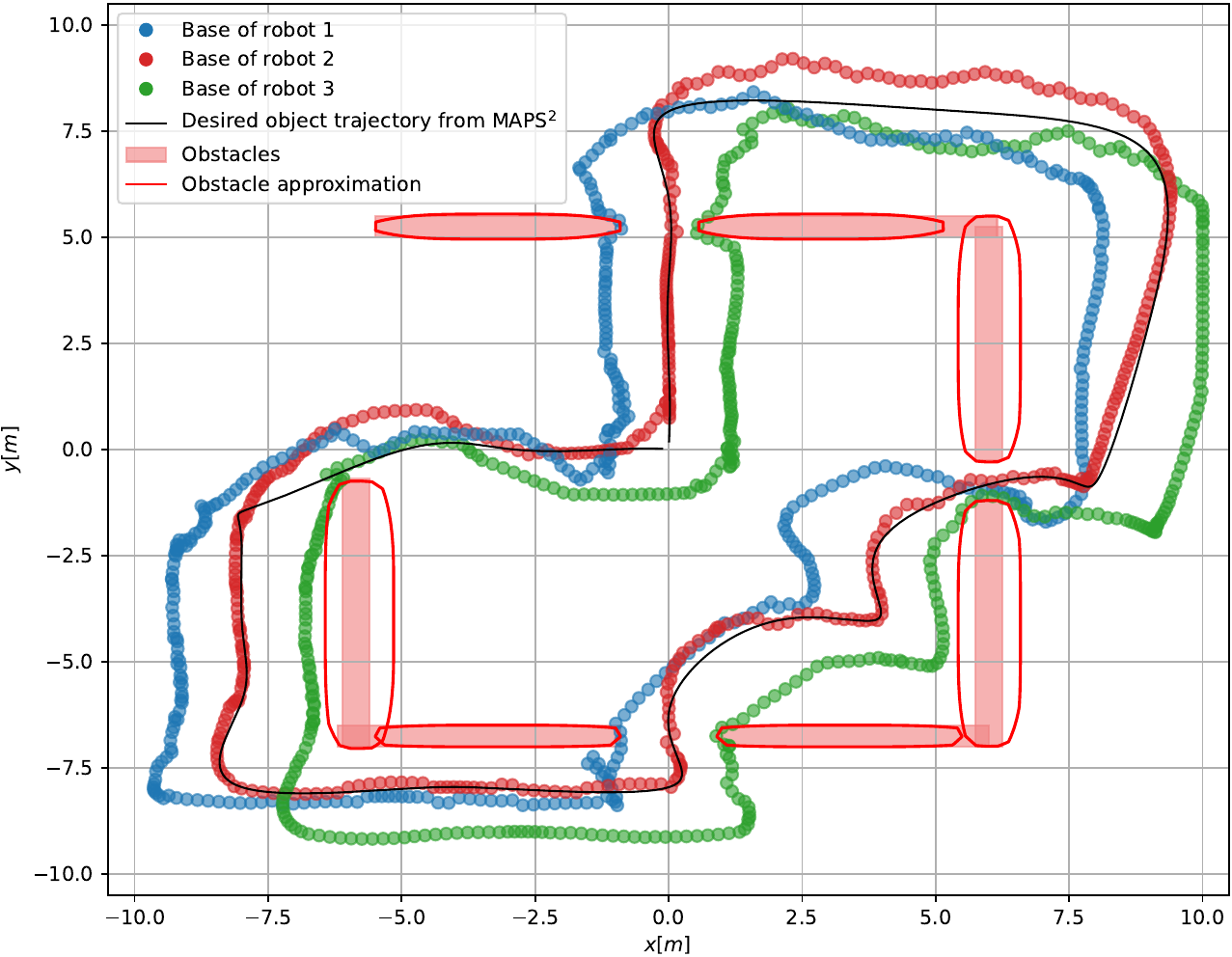}
    \end{subfigure}
    \hfill
    \begin{subfigure}[b]{0.49\textwidth}
        \includegraphics[height=5.8cm]{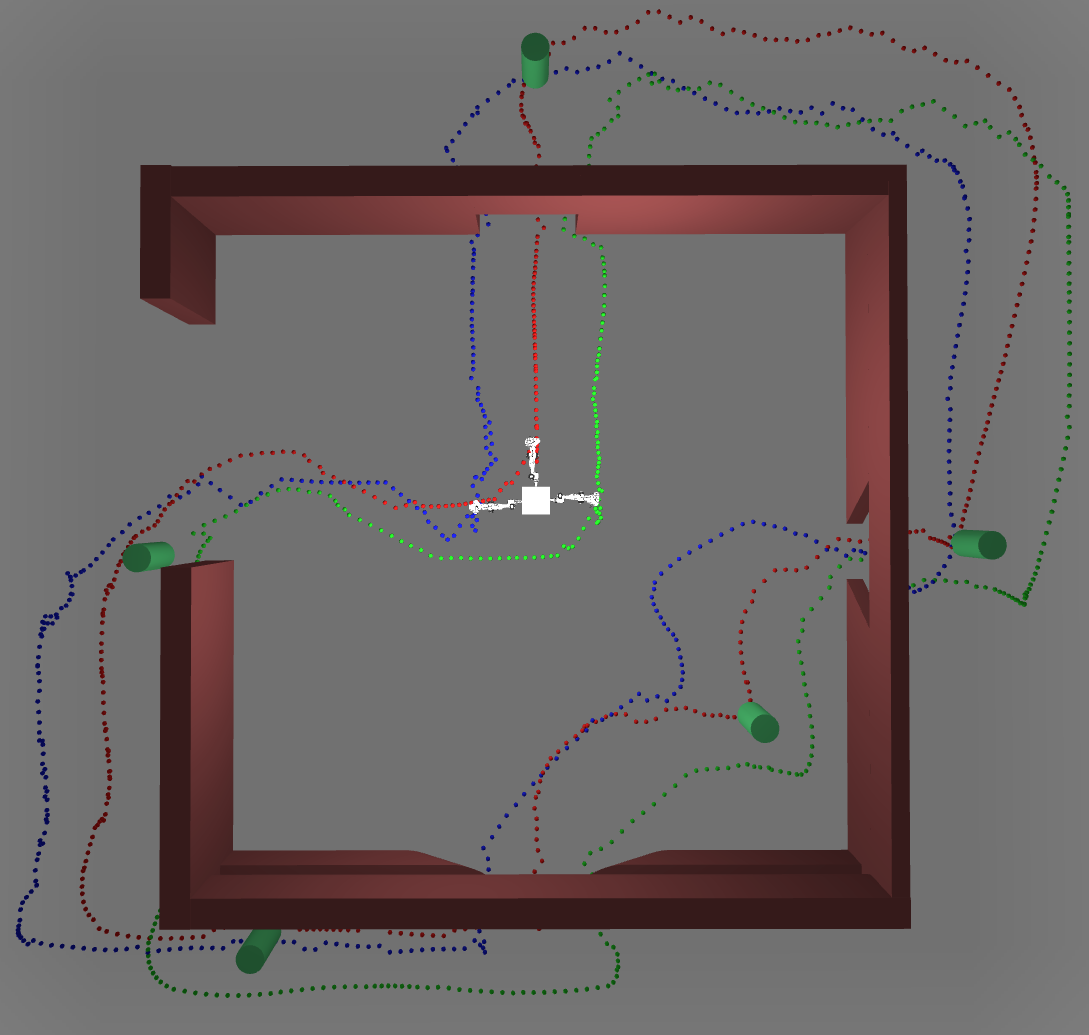}
    \end{subfigure}
       \caption{Desired trajectory from MAPS$^2$ along with the solution to \eqref{eq_passage:opt}}
       \label{fig_passage:footprint1}
\end{figure*}

\begin{figure*}[t]
    \centering
    \begin{subfigure}[b]{0.49\textwidth}
        \includegraphics[width=\linewidth]{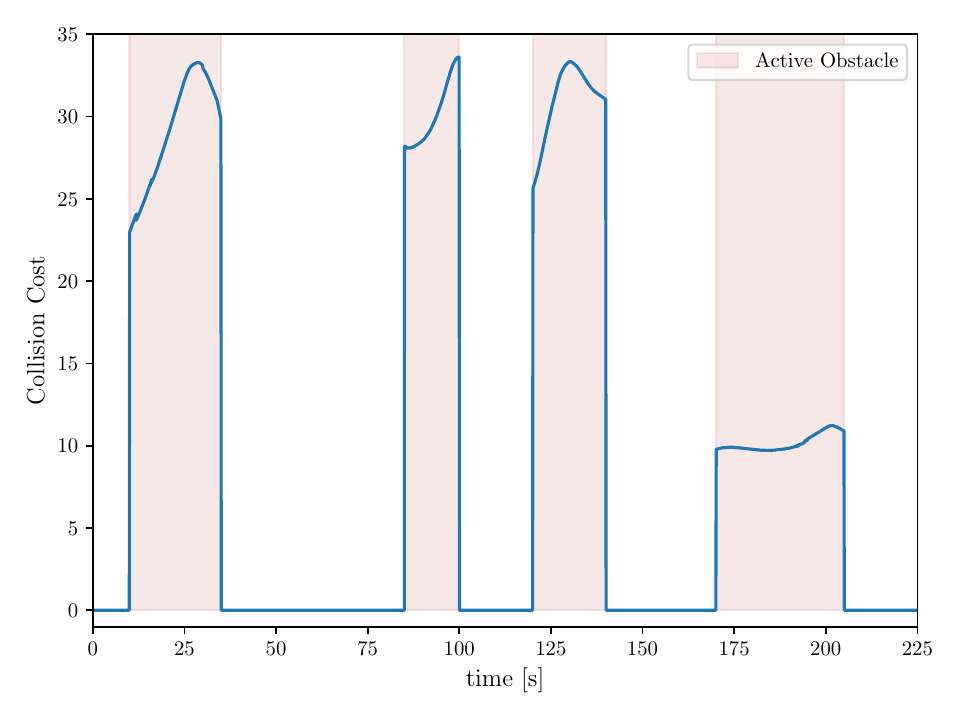}
        \caption{Collision cost over time.}
        \label{fig_trajopt:collision_cost}
    \end{subfigure}
    \hfill
    \begin{subfigure}[b]{0.49\textwidth}
        \includegraphics[width=\linewidth]{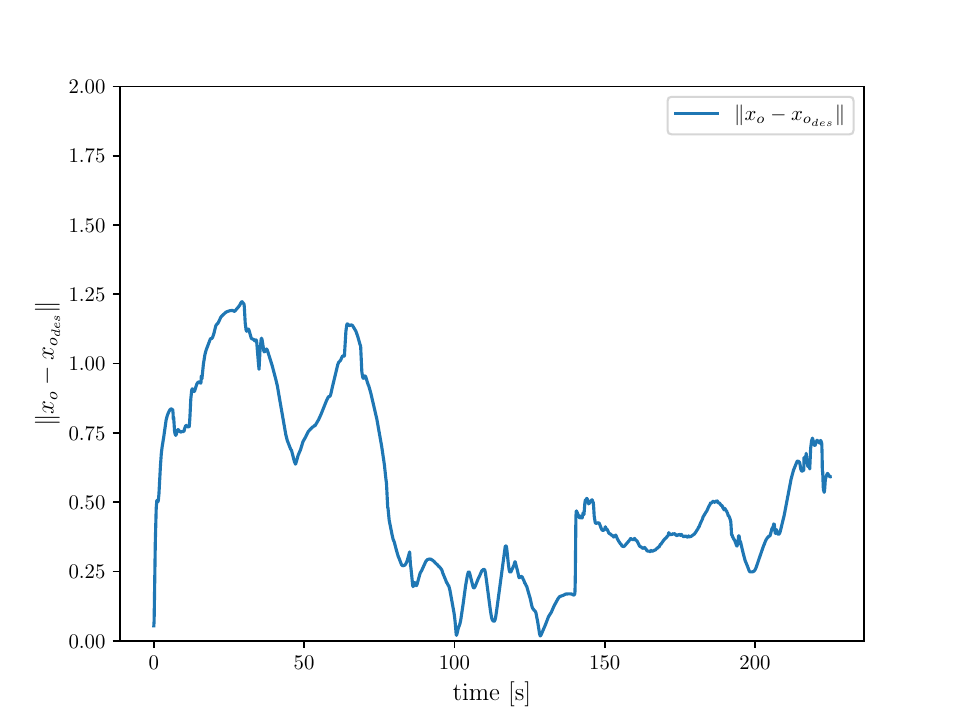}
        \caption{Object position error during execution.}
        \label{fig_trajopt:object_error}
    \end{subfigure}
    \vskip\baselineskip
    \begin{subfigure}[b]{0.49\textwidth}
        \includegraphics[width=\linewidth]{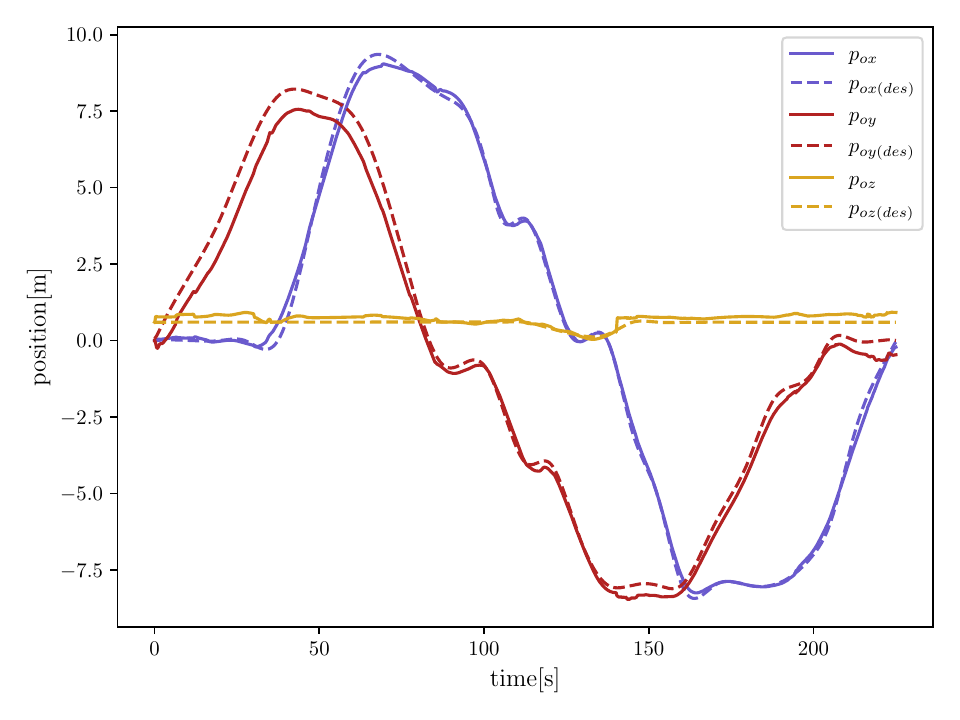}
        \caption{Object trajectory tracking.}
        \label{fig_trajopt:object_track}
    \end{subfigure}
    \hfill
    \begin{subfigure}[b]{0.49\textwidth}
        \includegraphics[width=\linewidth]{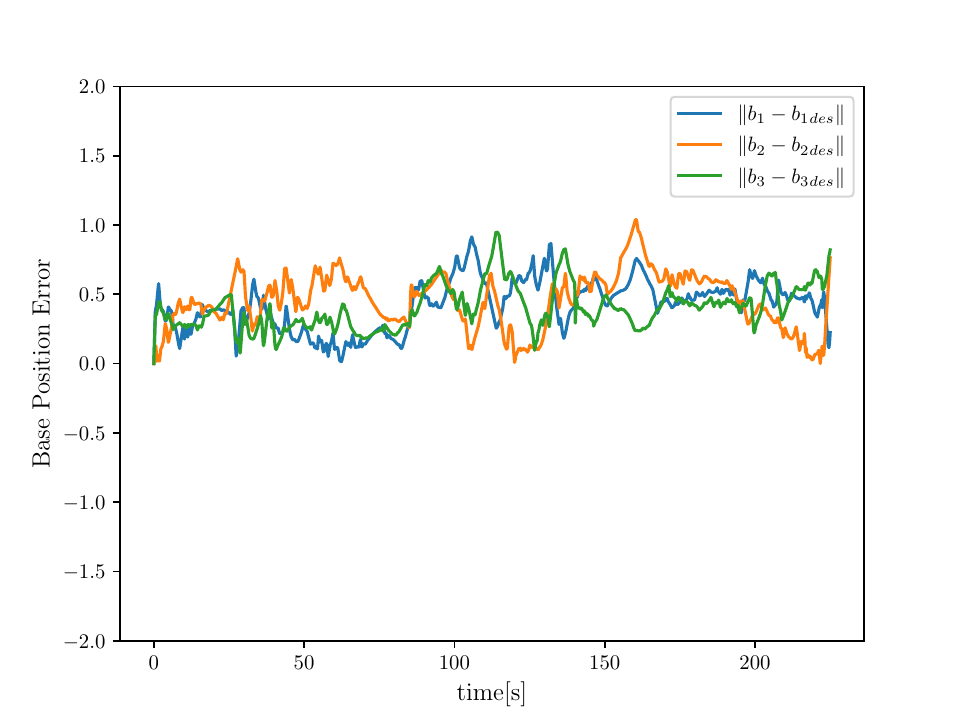}
        \caption{Base footprint tracking.}
        \label{fig_passage:footprinterror}
    \end{subfigure}
           \caption{Tracking errors of object and footprint over time}
           \label{fig_passage:signals}
\end{figure*}

\begin{figure*}[t]
    \centering
    \begin{subfigure}[b]{0.4\textwidth}
        \includegraphics[height=5cm]{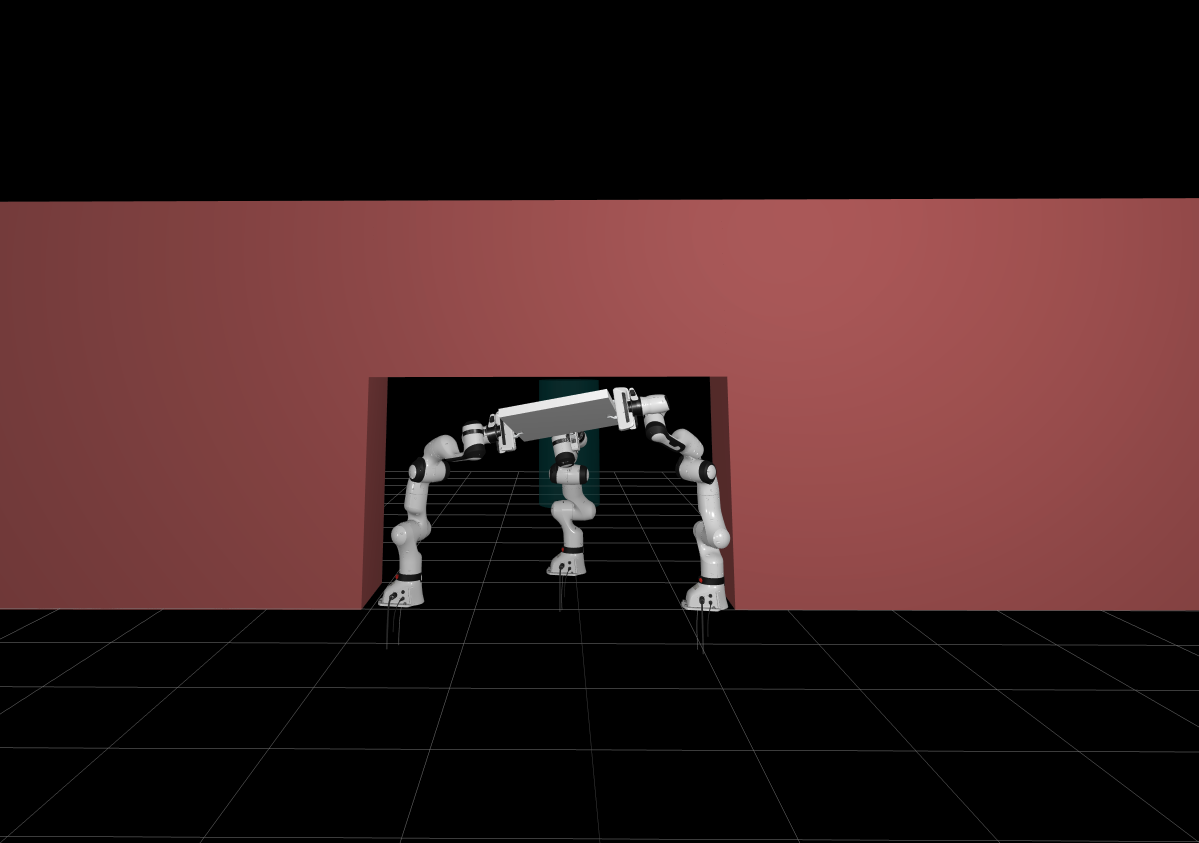}
        \caption{Passing through a square shape.}
        \label{fig_passage:square}
    \end{subfigure}
    \hfill
    \begin{subfigure}[b]{0.4\textwidth}
        \includegraphics[height=5cm]{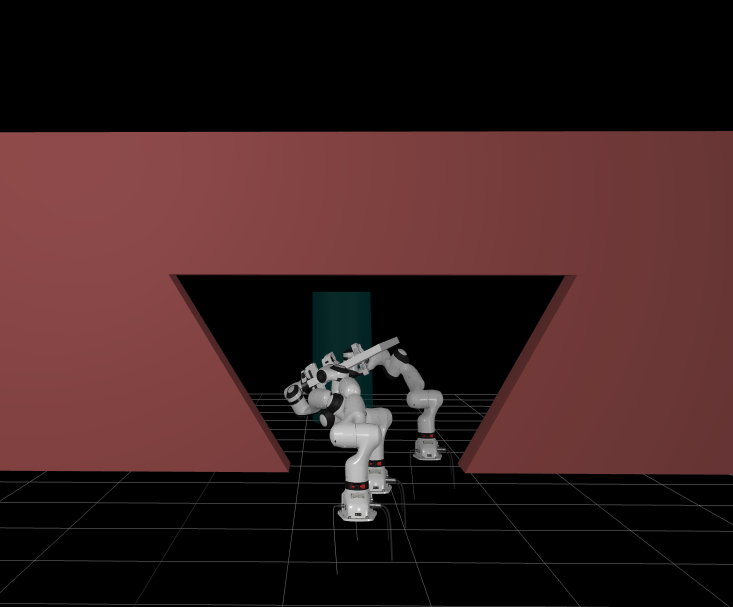}
        \caption{Passing through a triangle shape.}
        \label{fig_passage:triangle}
    \end{subfigure}
    \vskip\baselineskip
    \begin{subfigure}[b]{0.4\textwidth}
        \includegraphics[height=5.25cm]{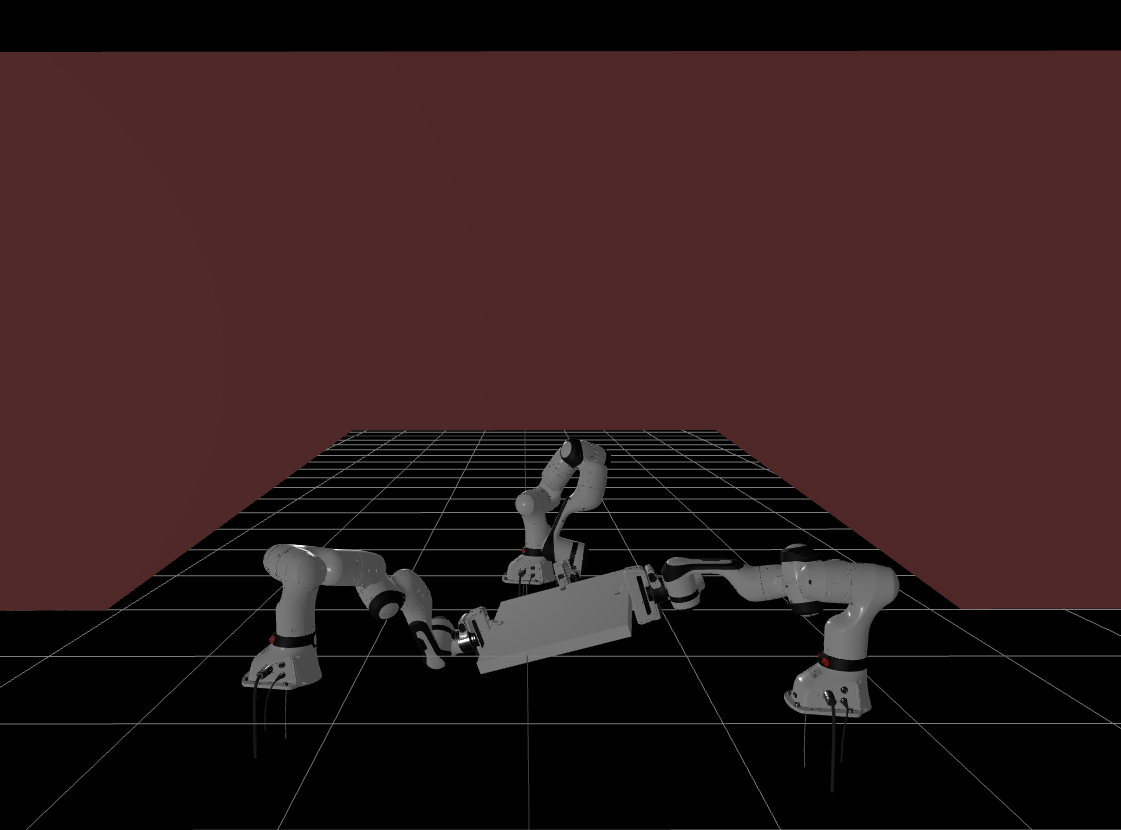}
        \caption{Passing through trapezoid shape.}
        \label{fig_passage:trapezoid}
    \end{subfigure}
    \hfill
    \begin{subfigure}[b]{0.4\textwidth}
        \includegraphics[height=5.25cm]{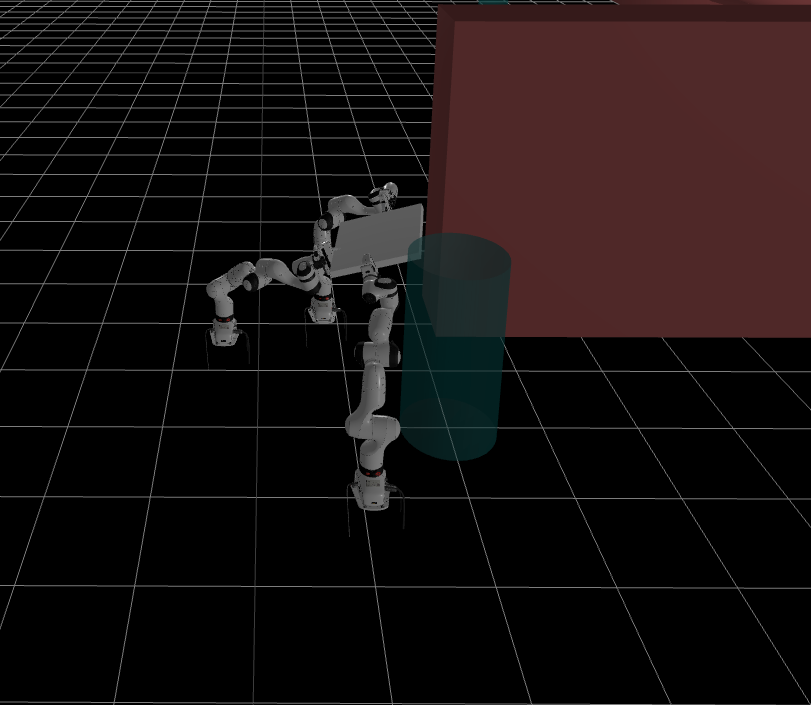}
        \caption{Avoiding obstacle.}
        \label{fig_passage:footprint}
    \end{subfigure}
    \caption{Snapshots of a simulation run.}
    \label{fig_passage:snap}
\end{figure*}

\section{Simulation Results}\label{sec:sim}

We evaluate the proposed framework on a system composed of three Panda mobile manipulators rigidly grasping a common object, as illustrated in Fig.~\ref{fig_passage:panda3}. Each manipulator is modelled in Drake~\cite{drake} as a floating-base system to emulate mobile capabilities. The workspace spans a $10\,[\mathrm{m}]\times 10\,[\mathrm{m}]$ area and contains multiple obstacles forming narrow passages of different shapes (square, triangular, and trapezoidal), shown in Fig.~\ref{fig_passage:ws}. The robots must transport the object through these passages while satisfying the given STL task specification.

\subsection{Task Specification}
The object must visit a sequence of waypoints with tolerance $\epsilon>0$ while maintaining a separation of at least $0.5\,\mathrm{m}$ from all obstacles, this translates into the following STL task:
\begin{align*}
    \varphi =\;&
    \mathcal{G}_{[37,38]}\big(\|x_o - [0\;\;8\;\;0.6]^\top\|\!\le\!\epsilon\big)\;\land
    \mathcal{G}_{[87,88]}\big(\|x_o - [8\;\;-1\;\;0.6]^\top\|\!\le\!\epsilon\big)\;\land \\
    &
    \mathcal{G}_{[112,113]}\big(\|x_o - [4\;\;-4\;\;0.6]^\top\|\!\le\!\epsilon\big)\;\land
    \mathcal{G}_{[150,151]}\big(\|x_o - [-5\;\;-8\;\;0.6]^\top\|\!\le\!\epsilon\big)\;\land \\
    &
    \mathcal{G}_{[200,201]}\big(\|x_o - [-7\;\;-1\;\;0.6]^\top\|\!\le\!\epsilon\big)\;\land
    \mathcal{F}_{[220,230]}\big(\|x_o - [0\;\;0\;\;0.6]^\top\|\!\le\!\epsilon\big)\;\land \\
    &
    \mathcal{G}_{[0,250]}\big(\|x_o - \mathrm{obs}\|\!\ge\!0.5\big).
\end{align*}
These waypoints are shown as green markers in Fig.~\ref{fig_passage:ws}. Their placement ensures that the robots must repeatedly enter and exit narrow corridors testing collision avoidance with varying degree.

\subsection{Modeling Considerations}
A number of practical considerations are required to faithfully simulate cooperative manipulation in high-fidelity software Drake:

\begin{itemize}
    \item {Rigid grasp modelling:} Physics engines generally do not support closed kinematic chains. We therefore approximate the rigid grasps using linear bushings, which introduce mild compliance while maintaining approximate rigidity.
    \item {Obstacle modelling in the Footprint Planner:} As discussed in Section \ref{footprint_planner}, each 2D obstacle is approximated by a super-ellipse. This avoids mixed-integer formulations that are incompatible with the solver SNOPT when used alongside nonlinear constraints. Figure \ref{fig_passage:footprint1} (left column) shows these approximations.
    \item {AutoDiff for collision penalties:} The collision cost $c_{\mathrm{collision}}$ requires gradients of signed distances between all robot and obstacle geometries. Drake only provides these gradients through AutoDiff models. Thus, the entire system (including three robots and the grasped object) is converted to AutoDiff to evaluate the smooth collision penalty.
    \item {Geometric simplification of robots:} For collision checking, the manipulators’ collision geometry is approximated using bounding boxes, as Drake currently does not provide signed distances for meshes or convex hulls coupled with AutoDiff. These boxes are sufficient for computing smooth distance-based penalties used inside the IK problem.
    \item {Selective activation of collision cost:} The collision penalty is activated only when the object enters the vicinity of an obstacle. This reduces unnecessary computation during long collision-free segments of the trajectory.
\end{itemize}

% \subsection{Execution Pipeline}
% The pipeline proceeds as follows:
% \begin{enumerate}
%     \item A globally feasible object trajectory satisfying~$\varphi$ is generated using \textit{MAPS$^2$}.
%     \item The \textit{Footprint Planner} computes collision-free base trajectories $b_{\mathrm{des}}$ consistent with kinematic reachability.
%     \item The \textit{Inverse Kinematics} module computes desired joint trajectories $q_{\mathrm{des}}$ at each time step.
%     \item A gravity-compensated PD controller tracks $q_{\mathrm{des}}$:
%     \[
%         \tau_i(t)
%         = -k_p\big(q_i(t) - q_{i\mathrm{des}}[k]\big)
%           - k_v \dot{q}_i(t)
%           + g_i(q_i(t)),
%         \qquad t\in[t_k,t_{k+1}),
%     \]
%     where $g_i(q_i)$ denotes the gravity compensation term.
% \end{enumerate}

\subsection{Results}
Figure~\ref{fig_passage:footprint1} shows the MAPS$^2$ object trajectory (top) and the resulting base trajectories from the \textit{Footprint Planner} (bottom), together with their 2D obstacle approximations. In all cases, the bases successfully respect the super-ellipsoidal constraints.

Figure~\ref{fig_passage:footprinterror} displays the error between the tracked base positions and their planned trajectories. A persistent but bounded error arises from the relaxation allowed in the IK problem, which balances grasp maintenance, obstacle avoidance, and joint-limit feasibility.

The collision cost evolution is illustrated in Fig.~\ref{fig_trajopt:collision_cost}. Elevated values occur only when the object passes near obstacles, matching the regions where the collision cost is active. Object tracking performance is shown in Figs.~\ref{fig_trajopt:object_error} and~\ref{fig_trajopt:object_track}, demonstrating that the manipulators successfully carry the object along the STL-specified path through all narrow passages.

Finally, Fig.~\ref{fig_passage:snap} presents snapshots of the robots navigating the square, triangular, and trapezoidal passages, as well as avoiding an obstacle. These visualisations confirm that the framework coordinates the three arms effectively through tight environments while maintaining the rigid grasp constraints and avoiding collisions. \footnote{The video of the simulations can be found here: \url{https://youtu.be/9W4wOzRl5dg}}

\section{Conclusion}\label{sec:conclusion}

This work proposed a hybrid, multi-rate control framework for cooperative manipulation under spatio-temporal task specifications and environmental constraints. The approach combines offline discrete planning of an STL-satisfying object trajectory and collision-free base footprints with online constrained inverse-kinematics updates and continuous-time feedback control. The resulting closed-loop system integrates discrete planning decisions with continuous nonlinear dynamics, enabling coordinated reconfiguration of multiple mobile manipulators in response to narrow passages and obstacles. High-fidelity simulations with three Panda mobile manipulators demonstrate robust trajectory tracking, collision avoidance, and successful execution of hybrid tasks in constrained environments.

\bibliographystyle{elsarticle-num} 
\bibliography{references}

\end{document}